%% file: gsw_aaai_submission_cameraready.tex
\documentclass[letterpaper]{article} 
\usepackage{aaai2026}  
\usepackage{times}  
\usepackage{helvet}  
\usepackage{courier}  
\usepackage[hyphens]{url}  
\usepackage{graphicx} 
\usepackage{natbib}  
\usepackage{caption} 
\usepackage{booktabs}
\usepackage{makecell}
\usepackage{amsfonts}       
\usepackage{nicefrac}       
\usepackage{microtype}      
\usepackage{tabularx} 
\usepackage{array}    
\usepackage{multirow}
\usepackage{amsmath}

\usepackage{enumitem}
\usepackage{amsthm} 

\usepackage{newfloat}
\usepackage{listings}
\DeclareCaptionStyle{ruled}{labelfont=normalfont,labelsep=colon,strut=off} 
\title{Beyond Fact Retrieval: Episodic Memory for RAG with Generative Semantic Workspaces}
\author {
    Shreyas Rajesh, Pavan Holur, Chenda Duan, David Chong, Vwani Roychowdhury
}
\affiliations{
    University of California, Los Angeles \\
    shreyasrajesh38@ucla.edu, pholur@ucla.edu, chenda@ucla.edu, \\
    davidchong13807@ucla.edu, vwani@ucla.edu
}

\begin{document}

\maketitle

\begin{abstract}
Large Language Models (LLMs) face fundamental challenges in long-context reasoning: many documents exceed their finite context windows, while performance on texts that do fit degrades with sequence length, necessitating their augmentation with external memory frameworks. Current solutions, which have evolved from retrieval using semantic embeddings to more sophisticated structured knowledge graphs representations for improved sense-making and associativity, are tailored for fact-based retrieval and fail to build the space-time-anchored narrative representations required for tracking entities through episodic events. To bridge this gap, we propose the \textbf{Generative Semantic Workspace} (GSW), a neuro-inspired generative memory framework that builds structured, interpretable representations of evolving situations, enabling LLMs to reason over evolving roles, actions, and spatiotemporal contexts. Our framework comprises an \textit{Operator}, which maps incoming observations to intermediate semantic structures, and a \textit{Reconciler}, which integrates these into a persistent workspace that enforces temporal, spatial, and logical coherence. On the Episodic Memory Benchmark (EpBench) \cite{huet_episodic_2025} comprising corpora ranging from 100k to 1M tokens in length, GSW outperforms existing RAG based baselines by up to \textbf{20\%}. Furthermore, GSW is highly efficient, reducing query-time context tokens by \textbf{51\%} compared to the next most token-efficient baseline, reducing inference time costs considerably. More broadly, GSW offers a concrete blueprint for endowing LLMs with human-like episodic memory, paving the way for more capable agents that can reason over long horizons. Code is available at \url{https://github.com/roychowdhuryresearch/gsw-memory}.
\end{abstract}

\section{Introduction}
\label{sec:introduction}
Large Language Models (LLMs) have transformed natural language understanding, but their ability to reason over long contexts is still limited by finite input windows. Even with token limits in the millions, large document collections can easily exceed these bounds. Performance can also degrade with context length due to phenomena like “context rot” and “lost-in-the-middle” effects \cite{liu_lost_2023, hong2025context}. A common workaround is Retrieval-Augmented Generation (RAG), which supplements the LLM’s input with only the most relevant retrieved content at query time. Standard RAG pipelines split documents into smaller chunks, encode them into dense embeddings, and retrieve the top-matching chunks based on semantic similarity to the query—allowing the LLM to focus on a relevant subset of the corpus during inference.

A key limitation of standard RAG methods is that each text chunk is embedded independently, which can lead to incomplete retrieval when a query depends on information spread across multiple chunks. Because similarity scores are computed in isolation, essential context may be missed. To address this, more recent approaches have adopted structured representations — such as knowledge graphs — that explicitly model relationships between chunks across the corpus. At query time, these graphs are traversed or queried to retrieve semantically connected chunks, enabling LLMs to perform more effective multi-hop reasoning and question answering \cite{gutierrez_hipporag_2025, JimenezGutierrez2025HippoRAG, Edge2025GraphRAG, Guo2024LightRAG}. 

These methods have primarily been evaluated on fact-rich documents such as Wikipedia pages \cite{yang2018hotpotqa, xanh2020_2wikimultihop, trivedi2022musiquemultihopquestionssinglehop}.   
Yet \textbf{the vast majority of texts that LLMs encounter are not lists of  facts but narratives of evolving real-world situations}.  Crime reports, political briefings, corporate filings, legislative records, war dispatches, and multi-day news coverage all describe \textbf{actors} (people, organizations, nations) that adopt \textbf{roles} (suspect, regulator, bidder, combatant) and transition through \textbf{states} (arrested → arraigned → released; startup → unicorn → acquired) while interacting across \textbf{space and time} \cite{goodreads, shahsavari2020corona, tangherlini2020bridgegate}. 



We contend that reasoning over such documents would be much more accurate and energy efficient, if one indexed the documents in terms of \textbf{an internal world model}— a structured representation that keeps track of \textit{who} is involved, \textit{what} was done, \textit{where} and \textit{when} events occur, \textit{how} roles change, and \textit{what} consequences follow.  Indeed, to achieve such a goal, humans possess \textit{episodic memory}
\cite{tulving_episodic_1972, tulving2002episodic} enabling us not only to plan and reason to seamlessly operate in the real world, but also to create new or update existing world models by reasoning across multiple experiences \cite{schacter2007remembering, hassabis2007deconstructing}. 

In this work, we introduce the \textbf{Generative Semantic Workspace} (GSW), a unifying computational framework for modeling world knowledge as structured, probabilistic semantics in the era of Large Language Models (LLMs). GSW formalizes how an intelligent agent—human or artificial—constructs and updates an internal representation of evolving situations from sequential input (e.g., text, video, or dialogue modalities). These representations are interpretable, actor-centric, and predictive: they reflect semantic regularities in the past while projecting likely future outcomes. GSW may be viewed as an instance of \textit{episodic memory} that can be integrated into LLM-based systems as a reasoning and memory module, serving as a symbolic bridge between language and latent world models.

To illustrate how GSW can help LLMs reason accurately,  we evaluate it on the Episodic Memory Benchmark (EpBench) \cite{huet_episodic_2025}, that has recently been introduced as a way to benchmark the episodic memory-like capabilities of LLMs. Following are excerpts from two different documents that relate to an entity, Carter Stewart, in this EpBench dataset: 
\begin{quote}
\textbf{Document \#1: }The imposing structure loomed before him, its grand facade a testament to both artistry and scientific achievement ...... As he stepped into the \textbf{Metropolitan Museum of Art}, the echoing chatter of excited voices ...... The antique clock in the main hall chimed, its resonant tones reminding him of the date: \textbf{September 22, 2026} .... found himself particularly engrossed during the third presentation, where \textbf{Carter Stewart} explained statistical analysis with a clarity that left the audience spellbound.
\end{quote}

\begin{quote}
\textbf{Document \#2: } The air crackled with tension as \textbf{Carter Stewart} stepped onto the pristine greens of \textbf{Bethpage Black Course} on \textbf{March 23, 2024} ...... Carter discussed implications of research, his fingers trembling slightly as he adjusted his microphone.
\end{quote}

An agent reading the narrative in the first document faces a fundamentally different challenge than traditional fact retrieval. It must understand that ``he'' refers to a nameless protagonist, who attended a scientific conference where Carter Stewart spoke. The narrator's spatial context (Metropolitan Museum of Art) and temporal context (September 22, 2026), are stated only indirectly and more importantly have to be also assigned to Carter Stewart who is a presenter. GSW is able to create such representations as part of its working memory construction task: ``Carter Stewart: \textbf{Role:} A presenter at a Scientific Conference; \textbf{Date:} September 22, 2026, morning session; \textbf{Location:} The Metropolitan Museum of Art, \textbf{Topic:} statistical
analysis; \textbf{Implements Used:} presentation boards and holographic projectors.'' The second document is more straightforward and GSW creates a memory trace such as: ``Carter Stewart; \textbf{Role:} a researcher and presenter; \textbf{Location:} Bethpage Black Course; \textbf{Date:} March 23, 2024,  \textbf{Did What?:} Presented his research findings at a Scientific Conference.'' 



When presented with a task such as ``List all the unique locations and dates where Carter Stewart made presentations at Scientific Conference events." a query resolution module (see Section \ref{sec:expsetup}) searches through the GSW constructed from all 200 documents and identifies entities mentioned in the query (e.g., Carter Stewart)  that match query's intent (e.g., a presenter at scientific conference; another entity named Carter Stewart whose role is that of a baker by profession would be ignored) and then returns just the relevant portion of its memory, as in the preceding paragraph. This results in highly targeted and short texts that an LLM has to reason through to provide an answer. In contrast, current structured RAG methods are designed to facilitate retrieval of either whole chunks or community-level summaries that have different levels of similarity to the entities and other phrases in the query. For example, for this query  GraphRAG's \cite{Edge2025GraphRAG} summarization missed that Carter Stewart was at the same location as the protagonist in Document \#1, and included irrelevant text chunks which led to a list that misses one location and hallucinates two erroneous locations. HippoRAG2 \cite{JimenezGutierrez2025HippoRAG} retrieves the full text of both the relevant documents, along with many other documents, overwhelming the LLM and leading it to hallucinate three erroneous locations. For a more detailed comparison, see Section~\ref{sec:limitations}, and Tables~\ref{tab:dataset_stats_revised}, \ref{tab:epbench_2000_short}, and \ref{tab:token_cost_comparison_final}. 


In the rest of this paper, we detail the GSW framework (\textbf{Section \ref{sec:approach}}) and present a rigorous evaluation on two versions of the EpBench benchmark (\textbf{Section \ref{sec:expsetup}}). The results demonstrate a significant improvement over existing methods. On the EpBench-200 corpus, GSW achieves a state-of-the-art F1-score of 0.85, outperforming strong structured RAG baselines. This advantage is particularly pronounced in the most demanding queries requiring synthesis across as many as 17 different documents, where GSW improves recall by up to \textbf{20\%} over the next best approach as detailed in Table \ref{Table:epbench_200_bootstrap}. Furthermore, GSW is  efficient, reducing the number of context tokens sent to the LLM by \textbf{51\%} compared to the most token-efficient baseline, drastically lowers inference costs and reducing the rate of hallucination in question answering (see Table \ref{tab:token_cost_comparison_final}). We further show that this powerful combination of accuracy and token efficiency holds at scale; on the EpBench-2000 corpus, a 10x larger dataset, GSW again achieves a state-of-the-art F1-score of 0.773, outperforming the best baseline by more than \textbf{15\%} on overall recall (Table \ref{tab:epbench_2000_short}), positioning GSW as a robust and scalable solution for equipping LLMs with effective episodic memory.

Results and discussions are summarized in \textbf{Section~\ref{sec:res-discussion}} and a review of related literature is presented in \textbf{Section~\ref{sec:related_work}}. Finally, limitations and future work are discussed in \textbf{Section~\ref{sec:limitations}}. The full version of the paper \footnote{The full version of this paper with a detailed technical appendix and clear visualization of the GSW using an example is available at \url{https://arxiv.org/abs/2511.07587}.} provides supporting evidence, including manual evaluations performed to validate the power of GSW's episodic memory capabilities.


\begin{figure*}[t]
    \centering
    \includegraphics[width=\textwidth]{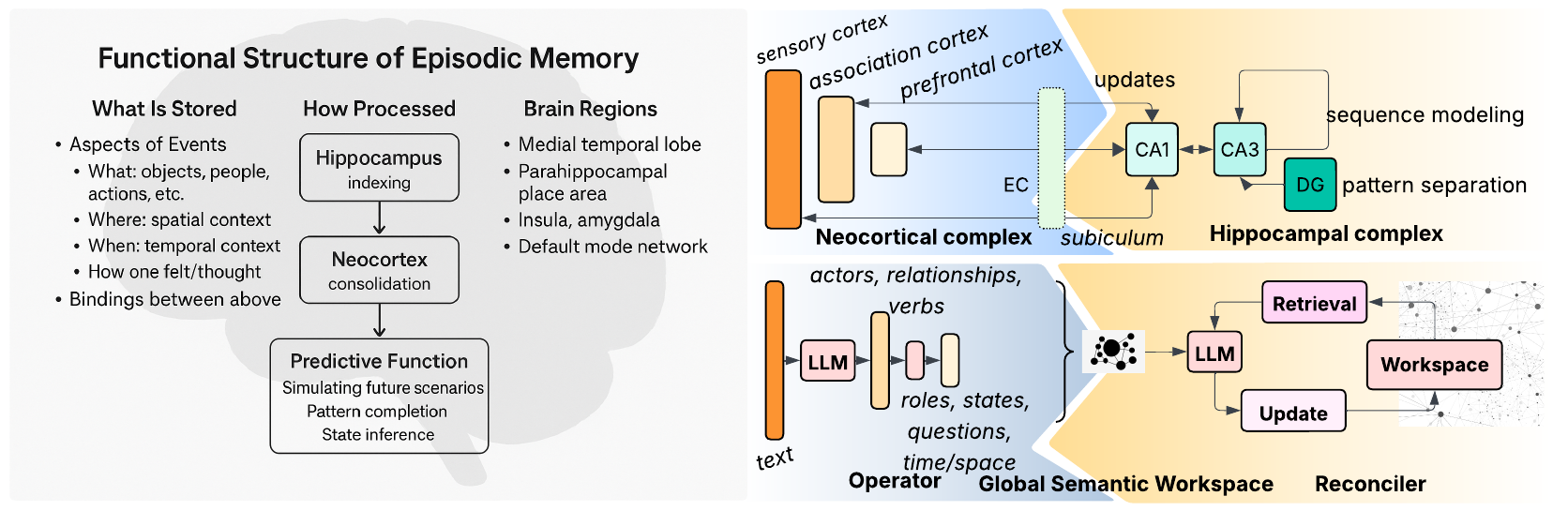}
    \caption{\textbf{Unifying Brain-Inspired and Generative Semantics for Episodic Memory Modeling} The hippocampal complex (DG, CA3, CA1) and neocortical regions (NC) inspire the \textit{Reconciler} (retrieval, workspace, update) and \textit{Operator} (LLM-driven semantic role extraction), respectively. The neocortical complex, responsible for context-rich consolidation and predictive modeling, aligns with the Operator module's functions. The hippocampal complex, which performs indexing, pattern separation, and sequence modeling, corresponds to the Reconciler. Together, the GSW framework offers a biologically inspired, interpretable model for simulating world knowledge from text inputs.
}
    \label{fig:motivation}
\end{figure*}

\section{The Generative Semantic Workspace (GSW) Framework}
\label{sec:approach}

In neuroscience, the neocortex is believed to encode hierarchical abstractions of entities, roles, and event templates \cite{george2009towards, Botvinik2008, felleman1991distributed}. The hippocampus, especially the CA3 module, plays a complementary role by binding these representations into coherent spatiotemporal sequences \cite{teyler1986hippocampal, rolls2013quantitative, eichenbaum2004hippocampus}. During sleep, this neocortical-hippocampal system engages in \textit{experience replay}, a process through which episodic traces are reactivated in reverse or forward order to consolidate memory and refine internal models \cite{olafsdottir2018role, louie2001temporally, wilson1994reactivation}. This back and forth supports both persistence and prediction of memory \cite{mcclelland1995there, rasch2013sleep}, key features of episodic memory. 

Motivated by this biological architecture (see Fig \ref{fig:motivation}), an effective memory framework requires a \textbf{structured representation} capable of encoding actors along with their evolving roles and states. Crucially, this representation must be capable of spatiotemporal grounding, linking entities and their interactions to specific times and locations, much like the binding function of the hippocampus. Finally, the framework must possess a process for \textbf{consolidating and updating these structures} as new information arrives, mirroring the way the neocortical-hippocampal loop constantly refines its world model.

\mbox{ \\ }
\noindent 
\textbf{From Episodic Memory to Generative Modeling of Situations and Narratives:} 
The central challenge, therefore, is to create a continuously evolving semantic model, which requires a bidirectional mapping between text and a structured representation. While early symbolic frameworks like PropBank~\cite{propbank} and FrameNet~\cite{framenet} attempted this, they were not designed for this full bidirectional process, relying instead on fixed ontologies that lacked the necessary probabilistic and dynamic interpretation.

\textit{LLMs now make this bidirectional mapping tractable.} They can both infer concise semantic identifiers from text and generate coherent narratives from those identifiers. This enables a new, efficient memory model where compact semantic traces are stored and reactivated in context. The formal model is presented next.

\subsection{A Probabilistic Model for Semantic Memory: The Operator Framework}
\label{subsec:Operator}

We now define a minimal schema for encoding these semantic elements—along with predictive cues, spatiotemporal attributes, and utilities—that serves as the foundation of the GSW framework for structured memory in LLMs. The agent must distill and maintain a semantic map from text to build a coherent semantic model.

To make this concrete, let's consider a single text input $C_n$ at some time step $n$ : \textit{Yesterday, in a swift response to a reported robbery, law enforcement officers apprehended Jonathan Miller, a 32-year-old resident of Greenview Avenue, in the downtown area.}

Explicit information in $C_n$ typically specifies a configuration of participating actors $a_1, \dots, a_K$ and the relations or interactions among them. The agent must distill and maintain a semantic map from these clues to build a coherent semantic model. Let's represent this interaction pattern at time step $n$ as (here each entry denotes an interaction from actor $a_i$ to $a_j$ as inferred from $C_n$):

$$
\mathcal{C}_n \approx \begin{pmatrix} 
(a_1 \rightarrow a_1)^n & \cdots & (a_1 \rightarrow a_K)^n \\
\vdots & \ddots & \vdots \\
(a_K \rightarrow a_1)^n & \cdots & (a_K \rightarrow a_K)^n \\
\end{pmatrix};
$$

\noindent\textbf{Actors, Roles and States}\\
The word `Miller`, in isolation, corresponds to a broad, unconditioned distribution over possible behaviors of a human. 
If `Miller` is likely to commit a crime, the agent would probably refer to Miller with a label `Criminal`. We call these labels \textit{roles}. \vspace*{0.5ex}
\\
\textbf{Role:} An identifier that specifies a distribution over potential actions that an actor $a_i \in \mathcal{A}$ may take toward other actors $a_j \in \mathcal{A}$:
\begin{equation}
\pi_r: \mathcal{A} \times \mathcal{A} \rightarrow [0, 1]
\end{equation}
where $\pi_r(a_i \rightarrow a_j)$ denotes the probability of $a_i$ acting on $a_j$ in role $r$.
For example, assigning the role of `criminal` to Miller increases the \textit{likelihood} that he will engage in actions such as \textit{committing a crime} against another actor or increasing the chances that Miller will \textit{attempt to flee} from `law enforcement`.

The agent would also \textit{know} that in addition to Miller being a \textit{criminal}, Miller has been \textit{caught}. Or perhaps he \textit{escaped}. We call these labels \textit{states}. \vspace*{0.5ex}\\
\textbf{State:} An identifier that induces a contextual attribute that modulates the probability distribution over actions available to an actor within a given role. Given an actor $a_i$ with role $r$, a state $s \in \mathcal{S}_r$ constrains the role-induced action distribution $\pi_r$:
\begin{equation}
\pi_{r,s}(a_i \rightarrow a_j) = \pi_r(a_i \rightarrow a_j \mid s),
\end{equation}
where $\pi_{r,s}$ denotes the subset of actions available to actor $a_i$ in state $s$. For instance, a \textit{criminal} in the state \textit{captured} may be limited to passive or compliant interactions, precluding actions such as fleeing or committing further crimes. Thus, states act as dynamic modifiers of an actor's interaction profile within a given situation.

\noindent\textbf{Verbs and Valences}\\
Verbs encode structured semantic attributes helping the agent to structure an event by drawing on prior experience, as verbs tend to generalize across contexts more reliably than nouns. They provide causal certificates for roles/states of actors. For example, understanding why Miller transitions from being \textit{free} to \textit{captured} relies on identifying the underlying interaction -- such as being arrested -- that bridges those states. A verb's valences are efficient means of capturing information needed for reasoning about future outcomes.  Verbs can be modeled similar to roles and states:
\begin{equation}
v(a_i \rightarrow a_j) : \mathcal{A} \times \mathcal{A} \rightarrow \mathcal{L}_v,
\end{equation}
where the valences $\ell_k \in \mathcal{L}_v$ signal the change in roles and states of the actors interacting via the verb. When Miller is running from the police, the \textit{next} state for Miller might be \textit{escaped} or \textit{caught}: a distribution of potential \textit{future} roles and states.

\noindent\textbf{Time and Space Continuity}\\
Spatiotemporal continuity constraints are crucial to capture world models, not only for individual actors but especially as interactions/verbs couple their coordinates. For instance, if Officers are actively apprehending Johnathan Miller in the Downtown area, then it enforces a shared location and time among the actors. Moreover, if the next day Miller is found in a city a thousand miles away, it would constrain his unobserved action to that of having flown and lead the agent to narrow down events that could have led to such a spatial shift. In effect, the flow of time and space regularizes the semantic map, biasing verb selection toward contextually coherent transitions.
If the position information derived from $\mathcal{C}_n$ at time step $n$ is $\mathcal{X}_n$ and the temporal information is $\mathcal{T}_n$, then: \vspace*{0.5ex}
\\
\textbf{Temporal continuity:} $\mathcal{T}_{n+1} - \mathcal{T}_n$ must be consistent with the expected temporal scope of $v$, \\
\textbf{Spatial proximity:} $\|\mathcal{X}_n(a_i) - \mathcal{X}_n(a_j)\|$ must fall within a valid range for the verb (e.g., \textit{tackle} requires physical closeness)

\noindent\textbf{Forward-Falling Questions to Capture Potential Outcomes and Actions }\\
The collection of roles/states, verbs, and spatiotemporal coordinates constrain the space of future progression and can be efficiently encoded as a set of questions $\mathcal{Q}_n$. For example, given that Miller has been arrested, ``When would Miller be indicted,'' ``where and when would the trial happen?'' ``Will he be free on bail?'' A prosecutor agent, for example, would need to start strategizing about such potential outcomes. 

A complete workspace instance can be written as a sampled distribution from an underlying ``Workspace'' generative process:
\begin{equation}
\mathcal{M}_n \sim p(\mathcal{A}, \mathcal{R}, \mathcal{S}, \mathcal{V}, \mathcal{T}, \mathcal{X}, \mathcal{Q} \mid \mathcal{C}_{0:n})
\end{equation}
where $\mathcal{M}_n \mapsto q(\mathcal{M}_{n+1} \mid \mathcal{M}_n)$ models the likelihood of generating the next workspace instance.

\subsection{Enabling Recursive Updates: A State Space Approach (The Reconciler Framework)}
\label{subsec:Reconciler}

Given a single text input $\mathcal{C}_0$, GSW models the workspace instance $\mathcal{M}_0$ as $P(\mathcal{M}_0 | \mathcal{C}_{0})$. We seek to compute: $P(\mathcal{M}_n | \mathcal{C}_{0:n})$. For $\mathcal{M}_1$, we introduce $\mathcal{W}_1$, an intermediate representation to decompose $P(\mathcal{M}_1|\mathcal{C}_0, \mathcal{C}_1)$ into parts:
\begin{align}
P(\mathcal{M}_1 &\mid \mathcal{C}_0, \mathcal{C}_1) \nonumber \\
&= \sum_{\mathcal{M}_0, \mathcal{W}_1} P(\mathcal{M}_1 \mid \mathcal{M}_0, \mathcal{W}_1) \nonumber \\
&\quad \times P(\mathcal{M}_0 \mid \mathcal{C}_0) P(\mathcal{W}_1 \mid \mathcal{C}_1)
\end{align}
Here, we assume conditional independence between the workspace state $\mathcal{M}_0$ and the intermediate representation $\mathcal{W}_1$ given the context sequence, such that:
\begin{align}
P(\mathcal{M}_0, \mathcal{W}_1 &\mid \mathcal{C}_0, \mathcal{C}_1) \nonumber \\
&= P(\mathcal{M}_0 \mid \mathcal{C}_0) P(\mathcal{W}_1 \mid \mathcal{C}_1)
\end{align}
where we define $\mathcal{W}_1$ to depend solely on the current context $\mathcal{C}_1$, and $\mathcal{M}_0$ solely on the initial context $\mathcal{C}_0$. For an arbitrary step $n$:
\begin{align}
P(\mathcal{M}_n | \mathcal{C}_{0:n}) &= \sum_{\mathcal{M}_{n-1}, \mathcal{W}_n} P(\mathcal{M}_n | \mathcal{M}_{n-1}, \mathcal{W}_n) \nonumber \\
&\quad \times P(\mathcal{M}_{n-1} | \mathcal{C}_{0:(n-1)}) P(\mathcal{W}_n | \mathcal{C}_n)
\end{align}
Estimating a workspace instance $\mathcal{M}_n$ involves learning parameterized models for three components: the transition model, the prior workspace, and the context-derived augmentation. The prior workspace $\mathcal{M}_{n-1}$ is recursively computed from previous steps. The augmentation step produces an intermediate representation of the current context $\mathcal{C}_n$. We refer to the model estimating this distribution as the \textbf{Operator}. The transition model uses a Markovian assumption to produce the updated workspace instance by reconciling existing workspace semantic maps with new semantic information. We refer to this module as the \textbf{Reconciler}. Together, the Operator and Reconciler implement a sequential inference mechanism where the Operator maps each new context $\mathcal{C}_n$ to an intermediate state $\mathcal{W}_n$, and the Reconciler performs a structured update $\mathcal{M}_{n-1} \rightarrow \mathcal{M}_n$.

\section{ Question Answering with GSW}\label{sec:expsetup}

\begin{figure*}[t]
    \centering
    \includegraphics[width=0.65\textwidth]{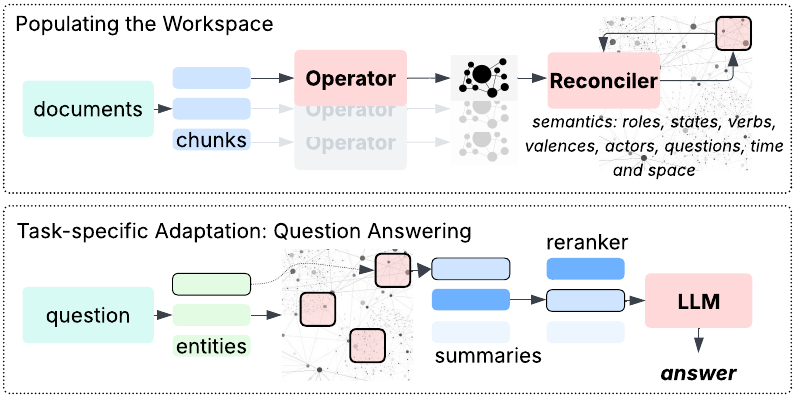}
    \caption{\textbf{Episodic Memory Creation and QA:} Figure illustrates the end-to-end process of constructing a workspace and question answering from the workspace. \textit{(top)} Large-scale text is segmented into semantically coherent chunks. Each chunk is processed by the \textit{Operator} model to generate a local workspace instance, represented as a semantic graph. These instances are incrementally integrated by the \textit{Reconciler} resulting in a unified Global Memory. \textit{(bottom)} During question answering, the system retrieves relevant portions of this memory by matching named entities in the query to identifiers in the semantic network. For each match, it reconstructs episodic summaries—contextual recreations of past situations—which are re-ranked and passed to an LLM to generate the final answer.}
    \label{fig:framework}
\end{figure*}

Figure~\ref{fig:framework} illustrates this process: memory construction via Operator and Reconciler modules, followed by retrieval, reranking and QA. As described in the caption, once a working memory instance is constructed, answering a query involves the following steps: the system first matches entities from the query to the GSW, then generates contextual summaries for those matched entities from the workspace, re-ranks the summaries for relevance, and finally passes the top-ranked summaries to an LLM to synthesize the answer.


\subsection{EpBench: An Episodic Memory Benchmark} \label{subsec:dataset}

Our experiments utilize the Episodic Memory Benchmark (EpBench) \cite{huet_episodic_2025}, a benchmark specifically designed to evaluate the capabilities of LLMs for episodic memory recall and reasoning over long narratives. Unlike many standard Question Answering (QA) benchmarks \cite{kovcisky2018narrativeqa, zhang_inftybench_2024, yang2018hotpotqa} -- focusing on localized factual retrieval -- EpBench targets core episodic capabilities: remembering specific events situated in unique spatiotemporal contexts and distinguishing between recurring events involving the same actors \cite{myside-yourside, insider-outsider}.

\input{Tables/data_content} 

EpBench documents are structured as synthetic books generated chapter-by-chapter from event templates (detailing date, location, entity, content) sampled from a larger universe, ensuring recurring elements that necessitate disambiguation and temporal tracking. Chapters are generated via LLM prompts and verified for coherence. Moreover, the same time/location/actors (collectively referred to as cues) appear across multiple chapters. For our evaluation, we use both the standard 200-chapter version and the extended 2000 chapter version of the dataset and report its Statistics in Table \ref{tab:dataset_stats_revised}.




\subsection{Evaluation Metrics}
\label{subsec:metrics}

To evaluate model performance on the EpBench dataset's queries (detailed in Section~\ref{subsec:dataset}), we adopt the LLM-as-a-Judge evaluation paradigm \cite{zheng_judging_2023}. For consistency, we strictly follow the LLM-based answer processing and extraction procedure outlined by the EpBench benchmark authors. This approach accounts for the possibility that model responses might be longer or more elaborate than the typically concise ground truth answers. These LLM extracted answers are then used to compute Precision, Recall and F1 scores which we report in Table \ref{Table:epbench_200_bootstrap}

\subsection{Baseline Methods}
\label{subsec:baselines}

We compare GSW against several baseline approaches: \textbf{Vanilla LLM}, standard \textbf{Embedding-based RAG} \cite{karpukhin-etal-2020-dense, ram_-context_2023} for which we utilized the \textbf{Voyage-03}\footnote{https://blog.voyageai.com/2024/09/18/voyage-3/} embedding model selected for its strong performance on retrieval benchmarks \cite{thakur2021beir} , and the structured RAG methods \textbf{GraphRAG} \cite{Edge2025GraphRAG}, \textbf{HippoRAG2}\cite{JimenezGutierrez2025HippoRAG}, and \textbf{LightRAG} \cite{Guo2024LightRAG}. 

\subsection{Implementation Details}

The GSW \textbf{Operator} (Section~\ref{subsec:Operator}) and \textbf{Reconciler} (Section~\ref{subsec:Reconciler}) were implemented by prompting GPT-4o~\cite{hurst2024gpt} according to task-specific instructions, using temperature set to 0 for deterministic behavior. To ensure fair comparison, we standardized both the maximum context utilization (limited to 17 chapters per query, matching the maximum relevant chapters per query) and the answer generation model (GPT-4o) across all evaluated methods. To generate an answer for a given query, we first identify named entities within the query text. These entities are then matched to corresponding nodes within the current GSW memory ($\mathcal{M}_{n}$) using simple string matching. Summaries for the matched entities -- aggregated from the GSW structure -- are then retrieved and re-ranked based on semantic similarity to the query. The final re-ranked summaries are provided to the LLM to answer the query as illustrated in Figure \ref{fig:framework}. 

\input{Tables/epbench_200_bootstrap}

\input{Tables/token_count} 

\section{Results and Discussion}
\label{sec:res-discussion}


\textbf{QA Performance:}
Table \ref{Table:epbench_200_bootstrap} presents a comparative analysis of GSW against the baseline methods  detailed in Section \ref{subsec:baselines} across Precision (P), Recall (R), and F1-Score (F1) metrics, categorized by the number of matching cues per query. Across the aggregated metrics, GSW achieves the highest overall F1-Score (0.850), Precision (0.865), and Recall (0.894), improving overall metrics by more than \textbf{10\%} over the next-best method. GSW also demonstrates consistent performance across the various Cue categories, achieving the highest score in \textbf{16 out of 18} individual metric computations, and ranking second in the remaining two, highlighting its robust performance across varying levels of episodic recall complexity.
Particularly noteworthy is GSW's performance in the `6+ Cues' category. \textit{This is the most demanding scenario}, where correct responses can require reasoning across information spanning up to 17 distinct chapters (see Table \ref{tab:dataset_stats_revised}). Even in this complex setting, GSW demonstrates robust efficacy and achieves the highest performance over all metrics: F1:0.834 P:0.891, R:0.822. In particular when compared to HippoRAG2, next most performant in this category, GSW outperforms it by approximately \textbf{20\%} in recall. \textit{Recall, in particular, measures a framework's ability to map queries to the correct chapter and context}, and it is revealing that for all competing frameworks recall decreases as the number of matching cues increases, whereas the GSW maintains consistently strong performance, highlighting the strength  of its structured representation in storing episodic information. Finally, the Vanilla LLM is consistently the poorest performing baseline (e.g overall F1 Score of 0.642) reaffirming the inherent difficulty of the episodic QA task and the necessity of specialized memory frameworks like the GSW.

\textbf{ Scalability on EpBench-2000}: To assess the scalability of our method, we evaluate GSW on the EpBench-2000 dataset, which increases the corpus size by 10 fold. 
 The results, presented in Table \ref{tab:epbench_2000_short}, show that GSW maintains its performance lead by achieving an overall F1-score of 0.773, which is \textbf{15\%   higher} than the strongest baseline (embedding RAG), and \textbf{22\% higher} than other structured RAG methods. Thus, GSW’s advantages in recall and reasoning persist even at a significantly larger scale. 
\input{Tables/epbench_2000_short}



\textbf{Token Efficiency:}
Beyond query performance, GSW demonstrates substantial improvements in token efficiency, as detailed in Table \ref{tab:token_cost_comparison_final}, which presents the average number of context tokens supplied to the LLM per query, and the corresponding cost for all compared methods. GSW achieves a remarkable \textbf{51\%} reduction in token usage when compared to the next most token-efficient baseline (GraphRAG). 
This advantage is even more pronounced when compared to stronger performing baselines such as Embedding RAG and HippoRAG2, against which GSW offers a token reduction of nearly \textbf{59\%}. 
GSW's efficient approach to query resolution contributes to the reduction in token count: Rather than passing entire chapters or raw document chunks, GSW utilizes its semantic structure to generate entity-specific summaries, thereby providing only targeted query-specific information to the LLM. This focused contextual information 
also reduces hallucinations as supported by the GSW's leading performance in the `0 Cues' category, where no matching cues are present in the source document.

Several additional \textbf{ablation studies} are presented in the full version, including the removal of identifier types (e.g., temporal and spatial tags), evaluations on a shortened version of the EpBench dataset, and comparisons across different retrieval strategies. These experiments highlight the contribution of each component in the GSW architecture and underscore the importance of principled memory querying. For qualitative insights into GSW's behavior and outputs, see the full version.

\section{Related Work}
\label{sec:related_work}
The relevant literature has been discussed in the Introduction, and a detailed literature review is included in the full version. While modern LLMs offer increasingly large context windows, processing quality degrades with extreme lengths \cite{leng2024long, hsieh_ruler_2024}, with performance notably dipping for information in the middle of long contexts \cite{liu_lost_2023}. This makes reliable episodic tracking challenging when relying solely on native context windows.

Retrieval-Augmented Generation (RAG) \cite{lewis_retrieval-augmented_2021, gao_retrieval-augmented_2024, karpukhin-etal-2020-dense} addresses this by retrieving relevant chunks using dense \cite{bert, reimers_sentence-bert_2019, lee_nv-embed_2025}, sparse \cite{robertson_probabilistic_2009}, or hybrid \cite{cormack2009reciprocal} embeddings. While effective for fact-based QA, standard RAG struggles to connect dispersed information due to chunk-based retrieval \cite{chen_walking_2023, merola2025reconstructing}. Structured approaches like GraphRAG \cite{Edge2025GraphRAG}, LightRAG \cite{Guo2024LightRAG} and HippoRAG \cite{gutierrez_hipporag_2025, JimenezGutierrez2025HippoRAG} mitigate this by modeling relationships and supporting multi-hop reasoning.

Other research efforts have targeted episodic memory more directly. Larimar \cite{das_larimar_2024} proposes modifications to the LLM's attention mechanism, while EM-LLM \cite{fountashuman} introduces memory components integrated with open-weight models. While promising, these approaches often require architectural changes or are designed for specific models. In contrast, GSW is a plug-and-play module compatible with any LLM, requiring no specialized training or fine-tuning.

\section{Concluding Remarks and Limitations}
\label{sec:limitations}

 In this work, we introduced the Generative Semantic Workspace (GSW) as a framework for equipping LLMs with human-like episodic memory. Its two core components—the Operator, which interprets local semantics within short context windows, and the Reconciler, which integrates and updates these representations over time—jointly construct a persistent, structured memory. This memory maps raw text into evolving configurations of roles, states, and interactions within a coherent global workspace. On the Episodic Memory Benchmark, GSW outperforms existing approaches in both accuracy and token efficiency, offering a scalable and interpretable alternative to long-context or retrieval-based systems.

Nevertheless, we identify key limitations and avenues for future work. Firstly, GSW's evaluation, while utilizing EpBench for its strengths in spatiotemporal assessment, is constrained by the limited scope of current episodic memory benchmarks in thoroughly probing the complex evolution of actor roles and states within extended narratives; we are developing a more comprehensive benchmark to address this gap. Secondly, the present GSW implementation relies on a strong closed-source LLM (GPT-4o). Empirical validation of promising open-source alternatives \cite{yang2024qwen2, grattafiori2024llama} within our framework is essential. Expanding GSW to diverse data modalities beyond text is also an important direction for future work.

\bibliography{custom}


\clearpage
\input{appendix}


\end{document}

%% file: Tables/data_content.tex

\begin{table}[htbp]
\centering
\small
\begin{tabular}{@{}ll@{}} 
\toprule
Statistic                               & Value \\
\midrule
Number of Chapters                      & 200 \\
Total Tokens                            & 102,870 \\
Total Queries (QA Pairs)                & 686 \\
Queries by Event Category               & \\ 
\quad (0 / 1 / 2 / 3-5 / 6+ Cues)     & 180 / 180 / 108 / 128 / 90 \\
\midrule
Max. Chapters Referenced per Query      & 17 \\
Min. Chapters Referenced per Query      & 0 \\
\bottomrule
\end{tabular}
\caption{\textbf{EpBench-200 Dataset Statistics.}}
\label{tab:dataset_stats_revised}
\end{table}

%% file: Tables/epbench_200_bootstrap.tex
\begin{table*}[t]
\centering
\scriptsize 
\setlength{\tabcolsep}{4pt} 
\begin{tabular}{@{}llcccccc@{}}
\toprule
\multirow{2}{*}{Metric} & \multirow{2}{*}{Method} & 0 Cues & 1 Cue & 2 Cues & 3-5 Cues & 6+ Cues & Overall \\
& & (N=180) & (N=180) & (N=108) & (N=128) & (N=90) & (N=686) \\
\midrule
\multirow{7}{*}{\textbf{P}}
& Vanilla LLM  & $0.840\pm0.019$ & $0.734\pm0.021$ & $0.735\pm0.026$ & $0.703\pm0.021$ & $0.806\pm0.028$ & $0.766\pm0.010$ \\
& Embedding RAG & $0.906\pm0.021$ & $\underline{0.745}\pm0.026$ & $0.803\pm0.028$ & $0.823\pm0.025$ & $0.886\pm0.029$ & $\underline{0.832}\pm0.012$ \\
& GraphRAG \cite{Edge2025GraphRAG} & $\underline{0.950}\pm0.016$ & $0.657\pm0.029$ & $0.677\pm0.034$ & $0.753\pm0.028$ & $0.816\pm0.035$ & $0.781\pm0.013$ \\
& HippoRAG2 \cite{JimenezGutierrez2025HippoRAG} & $0.829\pm0.027$ & $0.704\pm0.029$ & $\mathbf{0.817}\pm0.026$ & $\underline{0.839}\pm0.026$ & $\mathbf{0.940}\pm0.020$ & $0.812\pm0.013$ \\
& LightRAG \cite{Guo2024LightRAG} & $0.946\pm0.017$ & $0.668\pm0.029$ & $0.615\pm0.036$ & $0.695\pm0.031$ & $0.822\pm0.037$ & $0.763\pm0.014$ \\
\cmidrule(lr){2-8}
& GSW (Ours) & $\mathbf{0.978}\pm0.011$ & $\mathbf{0.755}\pm0.026$ & $\underline{0.810}\pm0.027$ & $\mathbf{0.878}\pm0.019$ & $\underline{0.890}\pm0.024$ & $\mathbf{0.865}\pm0.010$ \\
\midrule
\multirow{7}{*}{\textbf{R}}
& Vanilla LLM & $0.840\pm0.019$ & $0.781\pm0.021$ & $0.526\pm0.021$ & $0.419\pm0.017$ & $0.229\pm0.014$ & $0.616\pm0.011$ \\
& Embedding RAG & $0.906\pm0.021$ & $\underline{0.863}\pm0.025$ & $0.773\pm0.033$ & $0.746\pm0.027$ & $0.624\pm0.036$ & $\underline{0.807}\pm0.012$ \\
& GraphRAG \cite{Edge2025GraphRAG} & $\underline{0.950}\pm0.016$ & $0.764\pm0.031$ & $0.686\pm0.035$ & $0.645\pm0.026$ & $0.537\pm0.030$ & $0.748\pm0.014$ \\
& HippoRAG2 \cite{JimenezGutierrez2025HippoRAG} & $0.829\pm0.027$ & $0.823\pm0.026$ & $\underline{0.800}\pm0.029$ & $\underline{0.749}\pm0.026$ & $\underline{0.675}\pm0.030$ & $0.787\pm0.013$ \\
& LightRAG \cite{Guo2024LightRAG} & $0.946\pm0.017$ & $0.716\pm0.033$ & $0.628\pm0.035$ & $0.559\pm0.029$ & $0.458\pm0.029$ & $0.699\pm0.015$ \\
\cmidrule(lr){2-8}
& GSW (Ours) & $\mathbf{0.978}\pm0.011$ & $\mathbf{0.863}\pm0.025$ & $\mathbf{0.869}\pm0.023$ & $\mathbf{0.893}\pm0.015$ & $\mathbf{0.822}\pm0.022$ & $\mathbf{0.894}\pm0.009$ \\
\midrule
\multirow{6}{*}{\textbf{F1}}
& Vanilla LLM & $0.840\pm0.019$ & $0.709\pm0.022$ & $0.585\pm0.021$ & $0.476\pm0.017$ & $0.325\pm0.017$ & $0.629\pm0.010$ \\
& Embedding RAG & $0.906\pm0.021$ & $\underline{0.726}\pm0.026$ & $0.723\pm0.030$ & $0.745\pm0.026$ & $0.680\pm0.035$ & $\underline{0.771}\pm0.013$ \\
& GraphRAG \cite{Edge2025GraphRAG} & $\underline{0.950}\pm0.016$ & $0.625\pm0.029$ & $0.625\pm0.034$ & $0.657\pm0.026$ & $0.607\pm0.032$ & $0.714\pm0.013$ \\
& HippoRAG2 \cite{JimenezGutierrez2025HippoRAG} & $0.829\pm0.028$ & $0.676\pm0.028$ & $\underline{0.762}\pm0.028$ & $\underline{0.754}\pm0.025$ & $\underline{0.746}\pm0.027$ & $0.753\pm0.013$ \\
& LightRAG \cite{Guo2024LightRAG} & $0.946\pm0.017$ & $0.594\pm0.030$ & $0.587\pm0.032$ & $0.579\pm0.028$ & $0.561\pm0.030$ & $0.678\pm0.014$ \\
\cmidrule(lr){2-8}
& GSW (Ours) & $\mathbf{0.978}\pm0.011$ & $\mathbf{0.744}\pm0.026$ & $\mathbf{0.807}\pm0.024$ & $\mathbf{0.868}\pm0.016$ & $\mathbf{0.834}\pm0.022$ & $\mathbf{0.850}\pm0.010$ \\

\bottomrule
\end{tabular}
\caption{\textbf{GSW performance on Epbench-200 (200-Chapters Book)} Performance is grouped by metric (Precision, Recall, F1-Score) across different numbers of matching cues per query. (N=X) indicates questions per category. Error bars are estimated via bootstrap resampling. Best score in each column for each metric group is \textbf{bold}; second best is \underline{underlined}.}

\label{Table:epbench_200_bootstrap}
\end{table*}

%% file: Tables/token_count.tex

\begin{table}[htbp]
\centering
\small 
\setlength{\tabcolsep}{4pt} 
\begin{tabular}{@{}lrr@{}} 
\toprule
Method & \multicolumn{1}{c}{Avg. Tokens} & \multicolumn{1}{c}{Avg. Cost\footnotemark} \\ 
\midrule
Vanilla LLM                         & $\sim$101,120          & $\sim$\$0.2528 \\
Embedding RAG                       & $\sim$8,771            & $\sim$\$0.0219 \\
GraphRAG \cite{Edge2025GraphRAG}    & \underline{$\sim$7,340} & \underline{$\sim$\$0.0184} \\
HippoRAG2 \cite{JimenezGutierrez2025HippoRAG} & $\sim$8,771            & $\sim$\$0.0219 \\
LightRAG \cite{Guo2024LightRAG}     & $\sim$40,476           & $\sim$\$0.1012 \\
\midrule
GSW (Ours)                 & \textbf{$\sim$3,587}   & \textbf{$\sim$\$0.0090} \\
\bottomrule
\end{tabular}
\caption{\textbf{GSW's Efficiency}: Average context tokens passed to the LLM per query on EpBench-200, and the estimated cost to answer that query. GSW achieves the best performance (detailed in Table~\ref{Table:epbench_200_bootstrap}) with the significantly lowest token count and cost, as highlighted below. Best score in each column is \textbf{bold}; second best is \underline{underlined}.}
\label{tab:token_cost_comparison_final} 
\end{table}
\footnotetext{Cost calculated using GPT-4o pricing of \$2.50 per million tokens.}

%% file: Tables/epbench_2000_short.tex
\begin{table}[t]
\centering
\scriptsize
\setlength{\tabcolsep}{3pt} 
\renewcommand{\arraystretch}{1.05} 
\begin{tabular}{@{}lccc@{}}
\toprule
\textbf{Method} & \textbf{Precision} & \textbf{Recall} & \textbf{F1} \\
\midrule
Embedding RAG
  & $\underline{0.827}\,\pm\,0.014$ & $\underline{0.688}\,\pm\,0.015$ & $\underline{0.675}\,\pm\,0.015$ \\
GraphRAG
  & $0.761\,\pm\,0.017$ & $0.548\,\pm\,0.017$ & $0.544\,\pm\,0.017$ \\
HippoRAG2
  & $0.759\,\pm\,0.016$ & $0.648\,\pm\,0.016$ & $0.635\,\pm\,0.015$ \\
LightRAG
  & $0.649\,\pm\,0.018$ & $0.497\,\pm\,0.017$ & $0.494\,\pm\,0.016$ \\
\cmidrule(lr){1-4}
GSW (Ours) 
  & $\mathbf{0.830}\,\pm\,0.010$ & $\mathbf{0.796}\,\pm\,0.009$ & $\mathbf{0.773}\,\pm\,0.009$ \\
\bottomrule
\end{tabular}
\caption{\textbf{Overall performance on Epbench-2000 (2000-Chapters Book).} The same convention as in Table \ref{Table:epbench_200_bootstrap} is followed.}
\label{tab:epbench_2000_short}
\end{table}

%% file: appendix.tex
\section*{Appendix}

Our technical appendix is structured as follows:
\begin{enumerate}
    \item Appendix \ref{app:sec:llmprompts}: Prompts to LLM. 
    \item Appendix \ref{app:sec:gsw_example}: Example GSW Instance.
    \item Appendix \ref{app:sec:qa_example}: GSW QA Example. 
    \item Appendix \ref{app:sec:qualitative_analysis}: Qualitative Analysis of GSW performance.
    \item Appendix \ref{app:sec:exp}: Further Implementation Details.
    \item Appendix \ref{app:sec:ablation}: Ablation studies. 
    \item Appendix \ref{app:sec:relatedmemory}: Related work on Memory Augmentation for LLMs.
    \item Appendix \ref{app:sec:CompCost}: Computational Costs and Resources for Building the GSW.
    \item Appendix \ref{app:sec:relatedmodels}: Related Computational Models of Workspaces. 
\end{enumerate}

\clearpage

\setcounter{secnumdepth}{1}
\renewcommand{\thesection}{\Alph{section}}
\setcounter{section}{0}

\section{Prompts to the LLM}
\label{app:sec:llmprompts}
In the following section, we describe the prompts used by each component of our GSW framework. 

\subsection{Operator} 
\label{app:sec:operator_prompts}
\begin{figure*}[!hbtp]
\centering
\includegraphics[width=\linewidth]{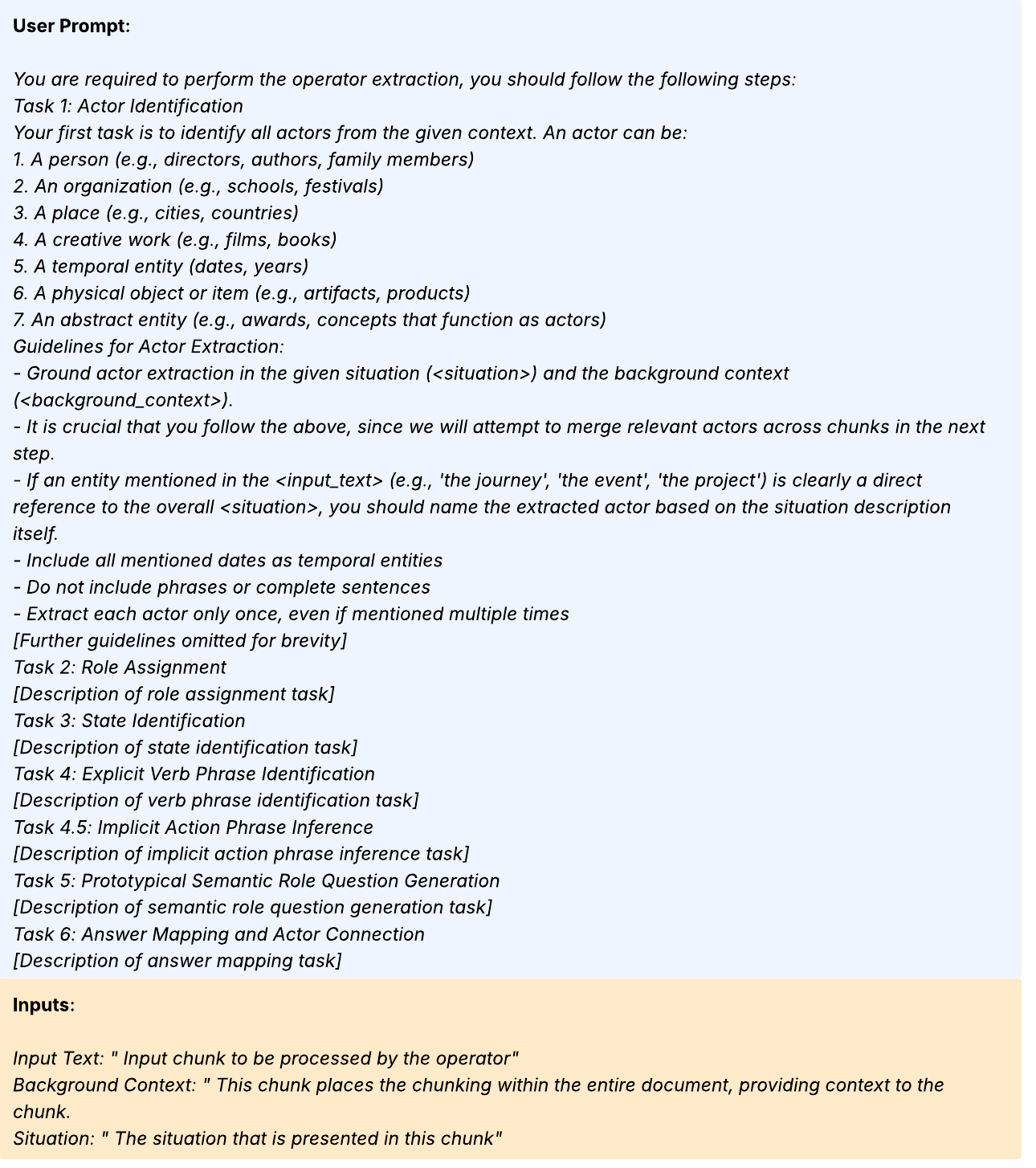}
\caption{LLM prompt for Operator extraction.\footnotemark}
\label{fig:prompt_operator}
\vspace{-1.5em}
\end{figure*}

\begin{figure*}[!hbtp]
\centering
\includegraphics[width=\linewidth]{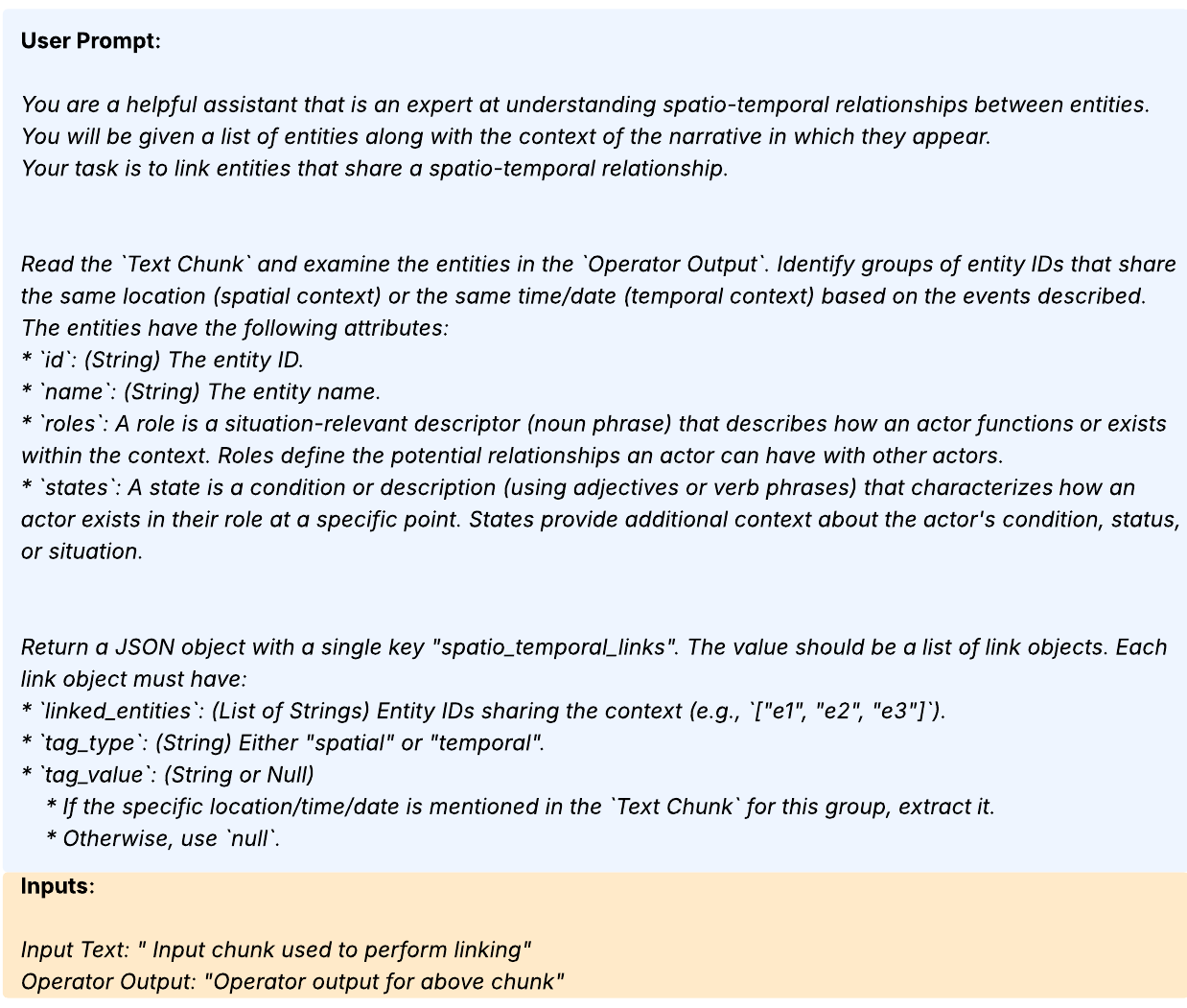}
\caption{
LLM prompt for Space Time coupling.
}
\label{fig:prompt_spacetime}
\vspace{-1.5em}
\end{figure*}

We present the prompt to generate operator representation in Fig \ref{fig:prompt_operator} and \ref{fig:prompt_spacetime}. The full prompt is considerably longer and includes detailed instructions for each task. For brevity, we have included the introduction and first task in full, with summaries of the remaining tasks. The complete prompt is available in our code repository.

\footnotetext{Background context generated according to contextual chunking by Anthropic, see \url{https://www.anthropic.com/news/contextual-retrieval}.}

\subsection{Reconciler}
\label{app:sec:reconciler_prompts}

\begin{figure*}[!hbtp]
\centering
\includegraphics[width=\linewidth]{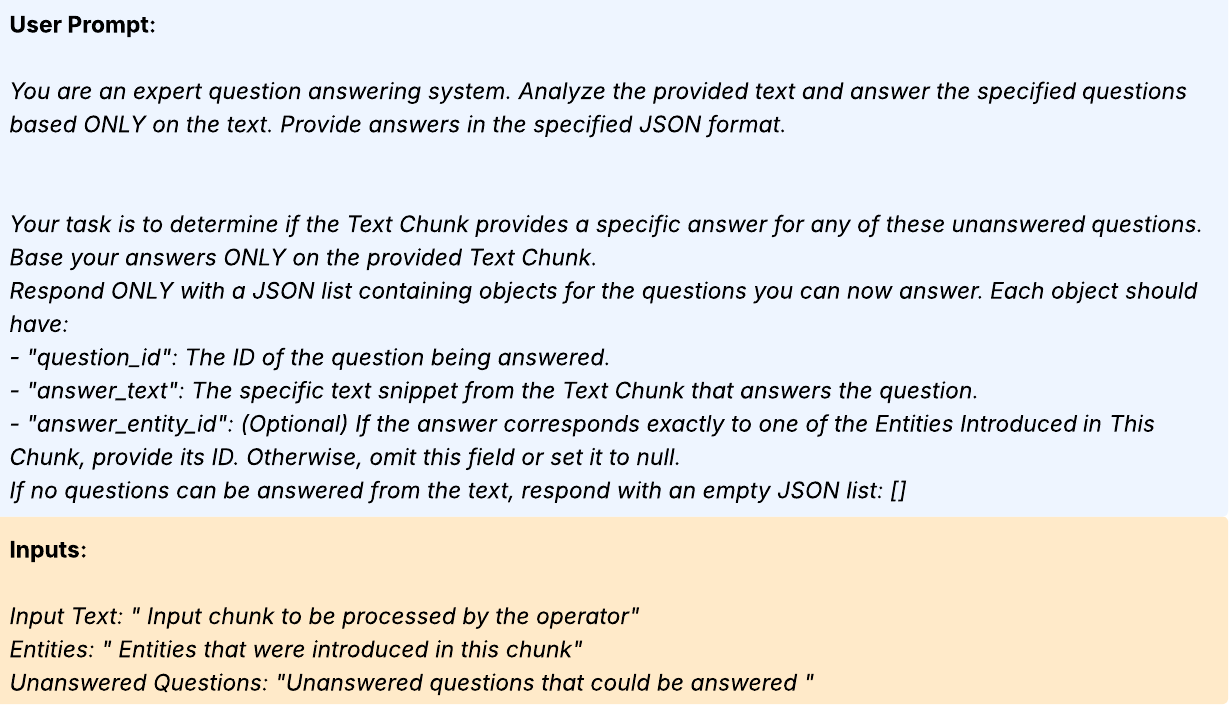}
\caption{
LLM prompt for QA reconciliation.
}
\label{fig:prompt_reconcile}
\vspace{-1.5em}
\end{figure*}

We present the prompt to reconcile unanswered queries with incoming context in Fig \ref{fig:prompt_reconcile}

\subsection{Question Answering}
\label{app:sec:qa_prompts}

\begin{figure*}[!t]
\centering
\includegraphics[width=0.8\textwidth]{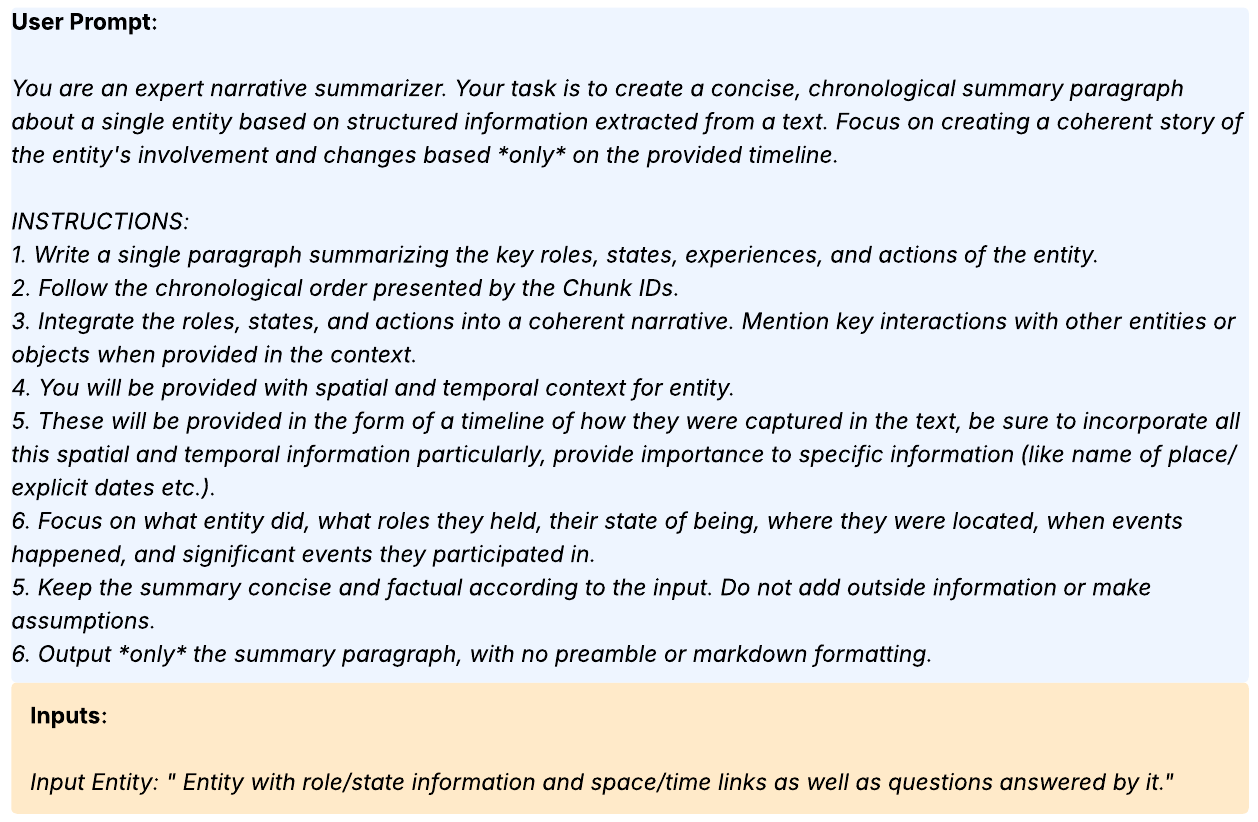}
\caption{
LLM prompt for entity summary generation.
}
\label{fig:prompt_summaries}
\vspace{-1.5em}
\end{figure*}

\begin{figure*}[hbtp]
\centering
\includegraphics[width=\linewidth]{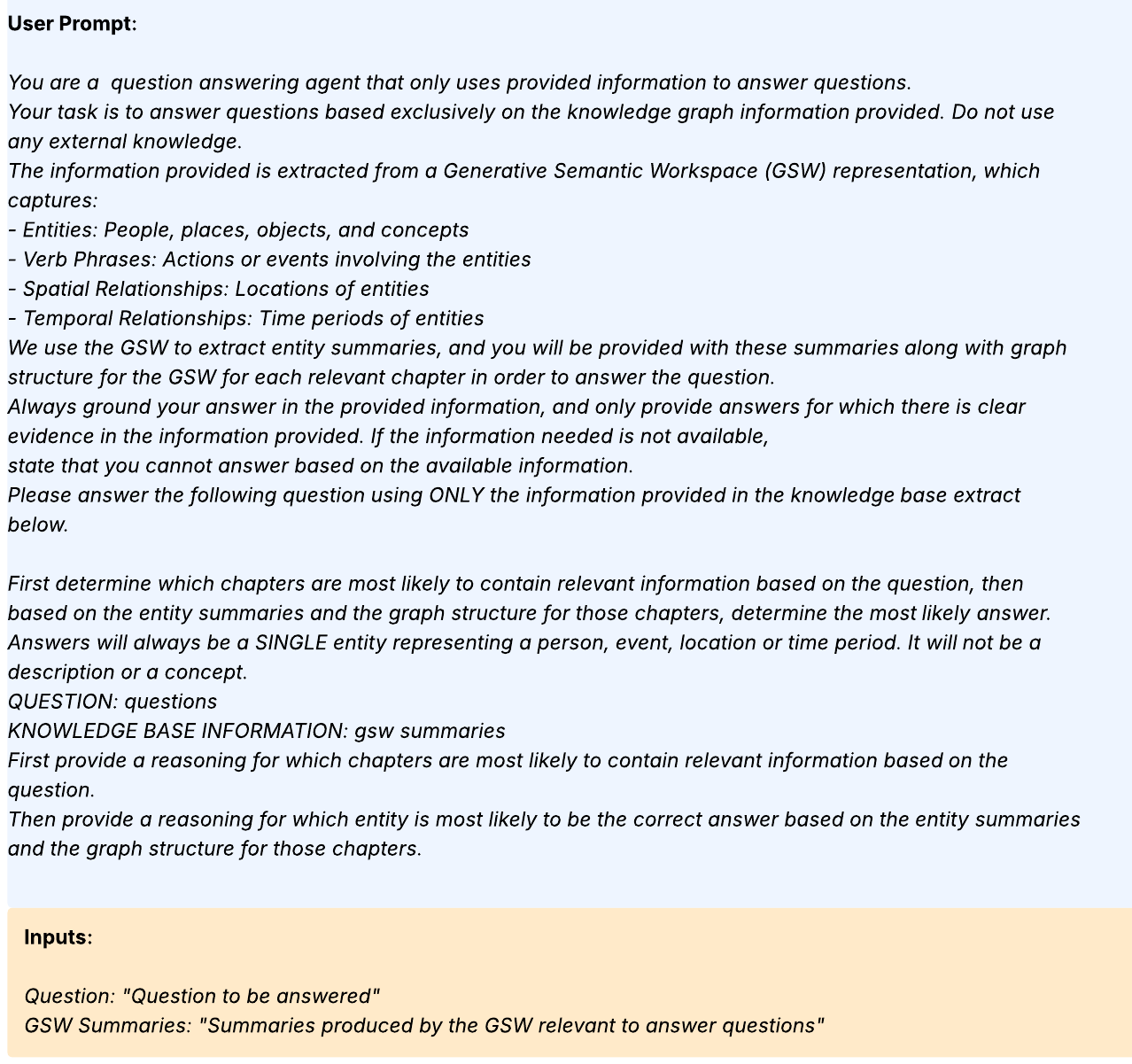}
\caption{
LLM prompt for final Question Answering.
}
\label{fig:prompt_answering}
\vspace{-1.5em}
\end{figure*}

We present the prompt to generate entity summaries which are passed to the answering agent in Fig \ref{fig:prompt_summaries} and the prompt used by the answering agent is presented in Fig \ref{fig:prompt_answering}

\section{Example GSW instance}
\label{app:sec:gsw_example}
\begin{figure*}[p]
\centering
\vspace*{-1cm}
\includegraphics[width=0.9\textwidth]{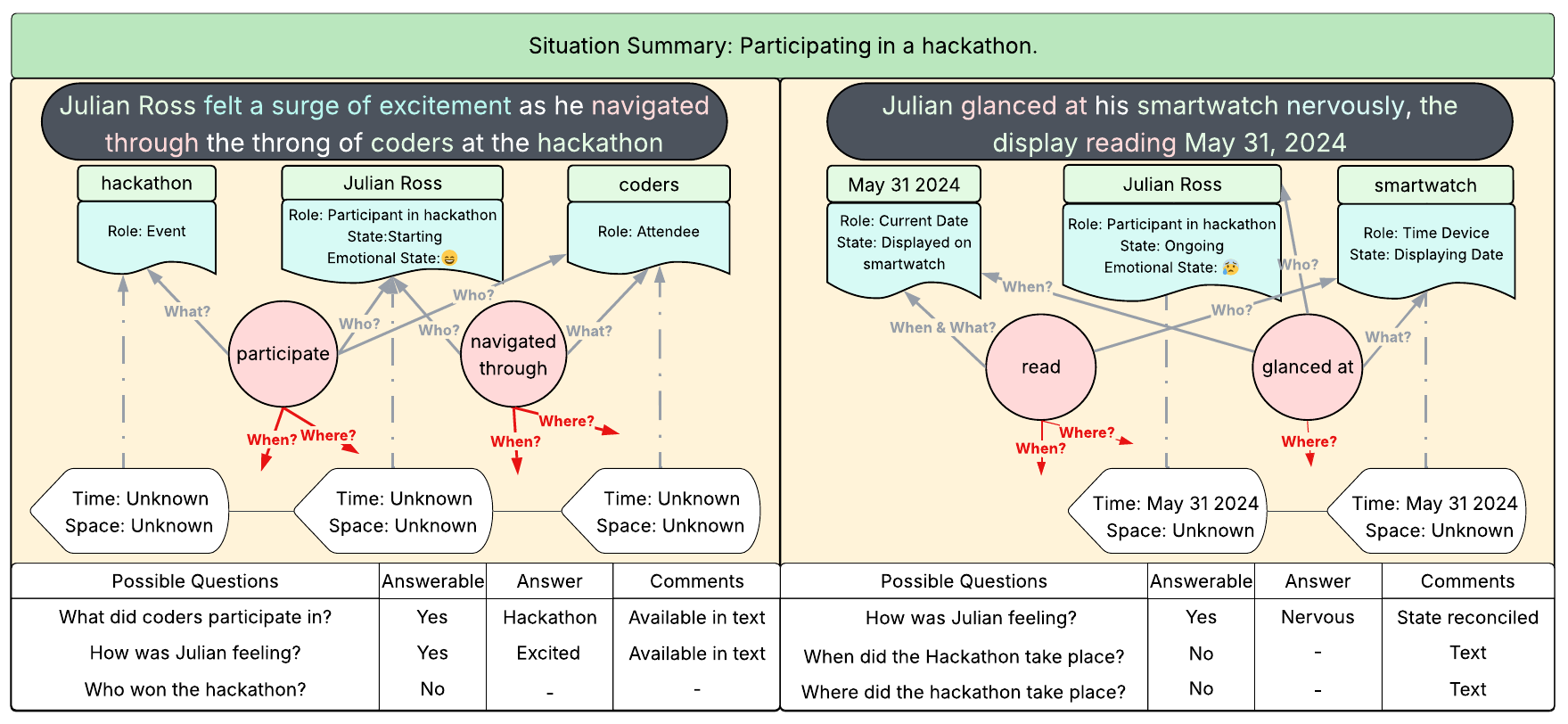}
\caption{\textbf{Operator example:} Operator instances of two different chunks, as the GSW framework processes a story.}
\label{app:fig:gsw_example1}
\vspace{1cm}

\includegraphics[width=0.9\textwidth]{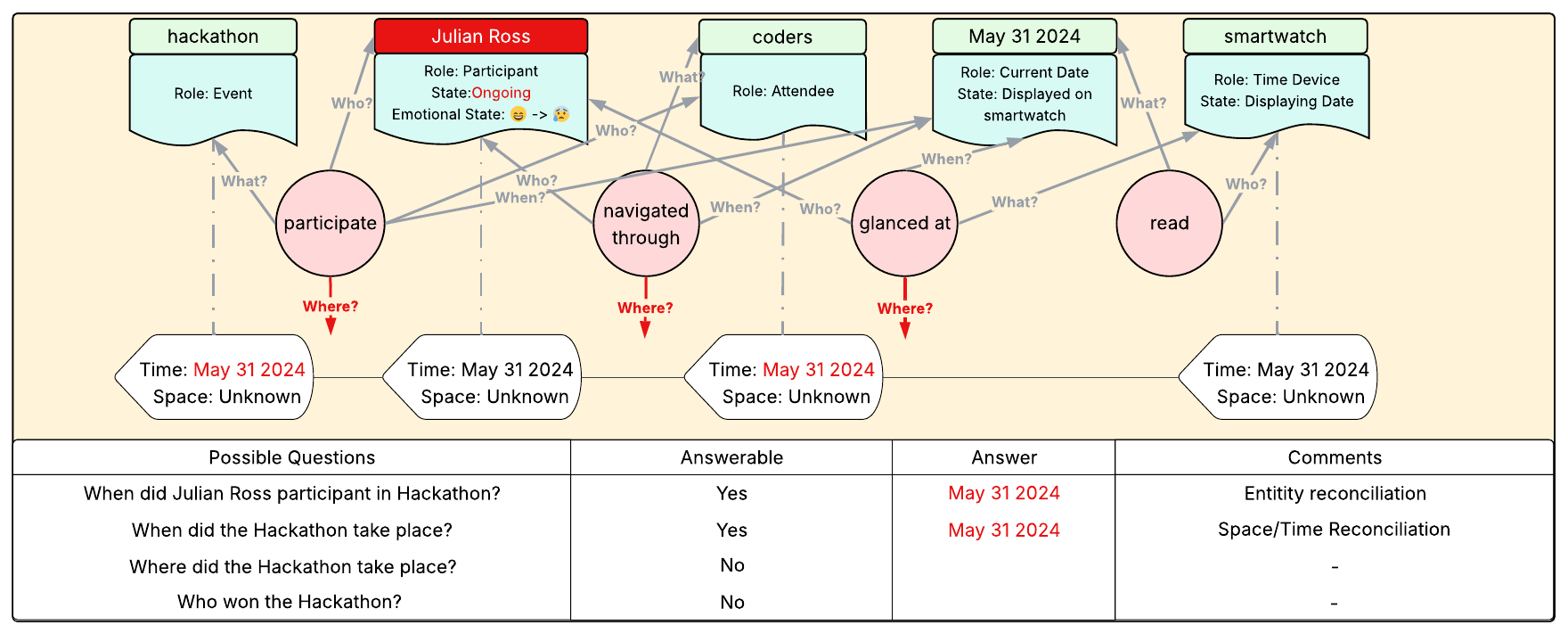}
\caption{\textbf{Reconciler example:} Reconciled result of the two chunks presented in Fig \ref{app:fig:gsw_example1}}
\label{app:fig:gsw_example2}
\vspace{1cm}

\includegraphics[width=0.9\textwidth]{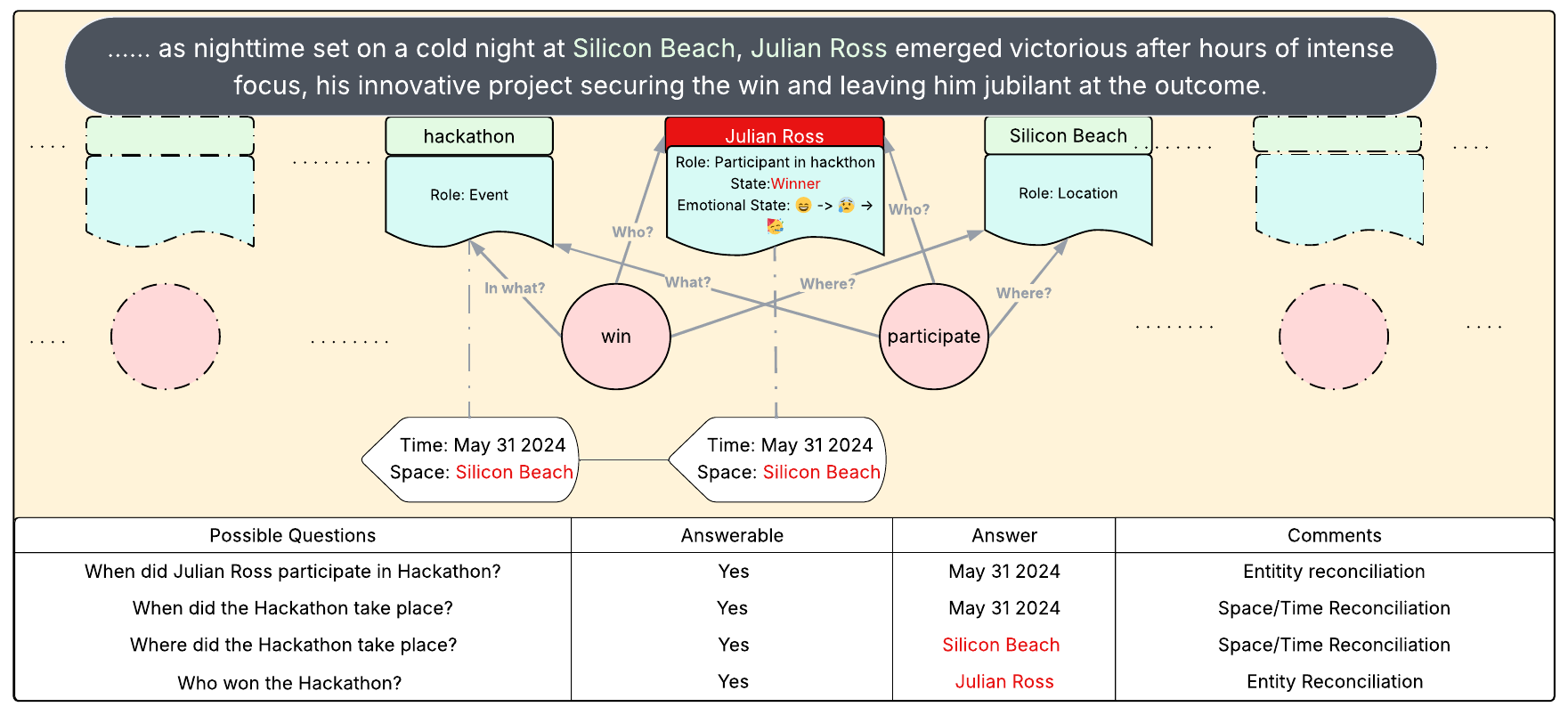}
\caption{\textbf{Final GSW:} A portion of the final reconciled GSW}
\label{app:fig:gsw_example3}
\end{figure*}

In this section we present an example instance of the GSW, highlighting the functionality of both the Operator and the Reconciler. Fig \ref{app:fig:gsw_example1} illustrates the operator representations for two separate chunks. Fig \ref{app:fig:gsw_example2} presents the result of reconciling the two representations presented in Fig \ref{app:fig:gsw_example1}. Finally Fig \ref{app:fig:gsw_example3} presents a portion of the final reconciled workspace with reconciled entities, space/time coupling and answered forward falling questions.

\section{GSW QA Example}
\label{app:sec:qa_example}
\begin{figure*}[!btp]
\centering
\includegraphics[width=\linewidth]{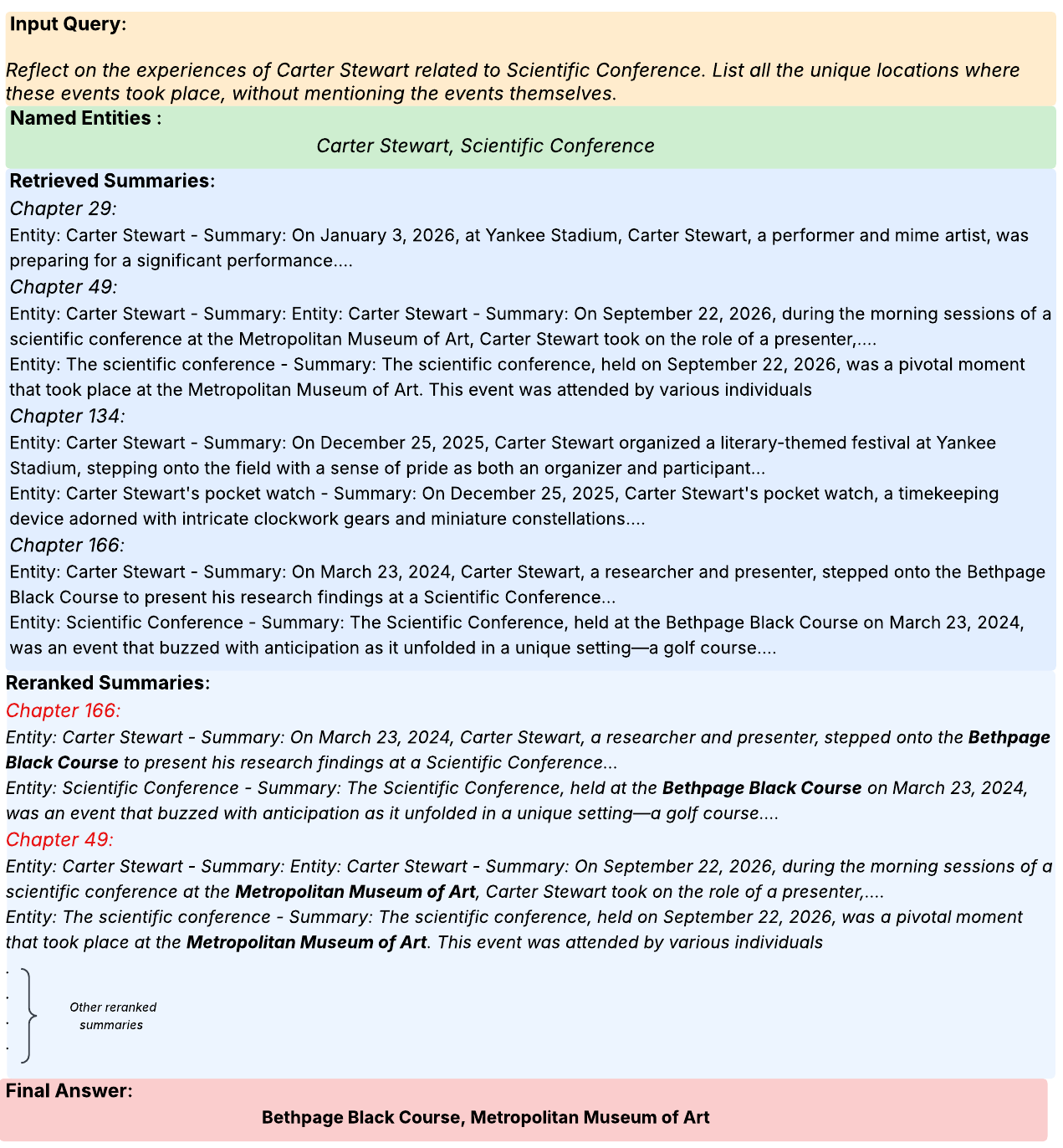}
\caption{
\textbf{Illustrative example of the GSW QA framework:} First, NER is performed on the input query to identify key entities. In this version of QA implementation these extracted entities are matched to the relevant GSW instances of chapters via string matching, and the entity-specific summaries (see Appendix \ref{app:sec:exp_qa}) from the GSWs are retrieved. Subsequently, these retrieved entity summaries are  re-ranked based on their semantic similarity to the input query—a score calculated using cosine similarity between their embeddings and the query's embedding. The figure displays a selection of initially retrieved summaries followed by the top re-ranked summaries. Finally, these re-ranked summaries are passed to an answering LLM, which then produces the final answer. As our considerably smaller average token count shows, our entity summaries are already concise, and only entity-relevant chapters are retrieved. Future implementations could leverage several avenues for further reduction in token counts without compromising performance. For example, in a query involving multiple entities, GSWs that have all the entities could be retrieved and sent to the LLM for a final answer; currently our re-ranking step ranks them at the top but we send summaries from other chapters as well, which is not necessary. 
}
\label{fig:qa_pipeline}
\vspace{-1.5em}
\end{figure*}

Figure \ref{fig:qa_pipeline} illustrates the end-to-end question answering (QA) pipeline of the GSW framework, showcasing how a sample query from the EpBench dataset is processed through each stage.

\section{Qualitative Analysis of GSW performance}
\label{app:sec:qualitative_analysis}
This section presents a qualitative analysis of selected queries to further illustrate GSW's superior performance and token efficiency compared to baseline methods, as detailed in Table \ref{tab:qual_analysis_comparison}. The chosen queries, whose full text and ground truth answers are provided in Table \ref{tab:qual_queries_answers}, are representative of varying complexity, with answers requiring the synthesis of information linked to two to seven distinct contextual cues. This detailed examination reveals specific failure modes in baseline approaches that GSW is naturally suited to overcome.

For instance, GraphRAG, which generates summaries of varying detail from source documents, frequently struggles with information loss and often provides an excessive volume of irrelevant context to the LLM, increasing the likelihood of hallucinations. This limitation is particularly noticeable in its handling of queries Q3 and Q4 (see Table \ref{tab:qual_analysis_comparison}). These queries demand precise spatial and temporal understanding of events, aspects that GraphRAG's summarization process does not natively or consistently capture, leading to missing information or inaccuracies in its responses.

HippoRAG2, on the other hand, processes every query through its knowledge graph --constructed by connecting semantically similar phrases across triples derived from all the chapters-- to identify the relevant chapters, and then provides full texts of these chapters as context to the LLM for a final answer. The strength of this approach is that they do not need to perform fine-grained analysis of the text -for example for dates and locations;- as long as their retrieval process identifies the right chapter, the onus is on the LLM to retrieve the relevant spatio-temporal information. This is an effective approach if the documents themselves are short and the number of documents needed to answer a query are few. In the EpBench data set the document size is around 500 tokens and the number of documents needed to answer some of the questions is 17; since QA cannot know the number of documents needed for any given question, 17 documents (chapters) were sent for each query across all evaluated methods.  As observed for queries Q2 and Q4 in Table \ref{tab:qual_analysis_comparison}, this strategy of providing full documents can overwhelm the LLM, leading to hallucinations or the failure to pinpoint the correct answer even when the right document with the necessary information is present in the retrieved context. Furthermore, there were instances (e.g., Q3, Q5) where HippoRAG2 failed to retrieve all the pertinent documents required to comprehensively answer the query. 

In contrast, GSW's structured representation and targeted summary generation (as detailed in Table \ref{tab:qual_analysis_comparison} showing 'None' for errors and lower token counts) effectively mitigate these issues. The ability of our GSW framework to collate and then structure spatio-temporal information scattered across the length of a document (via reconciliation) is aptly captured in the entity-level summary for Carter Stewart that is retrieved in response to Q2 (first three sentences are shown below):

\begin{quote}
    On September 22, 2026, during the morning sessions of a scientific conference at the Metropolitan Museum of Art, Carter Stewart took on the role of a presenter, delivering a final presentation that included statistical analysis using presentation boards and holographic projectors.
\end{quote}
The necessary information -- Carter Stewart, location, and time --in the original document came from three different   paragraphs; in fact, Carter Stewart is referred to as "He" until after location and time information is given: 
\begin{quote}
The imposing structure loomed before him, its grand facade a testament to both artistry and scientific achievement ...... As he stepped into the \textbf{Metropolitan Museum of Art}, the echoing chatter of excited voices ...... The antique clock in the main hall chimed, its resonant tones reminding him of the date: \textbf{September 22, 2026} ....  found himself particularly engrossed during the third presentation, where \textbf{Carter Stewart} explained statistical analysis with a clarity that left the audience spellbound."

\end{quote}
\input{Tables/app_sectionD_Queries}

\input{Tables/app_sectionD_perf}

\section{Implementation Details}
\label{app:sec:exp}
In this section, we provide further implementation details for the GSW as well as baselines implemented. 

\subsection{Operator}
\label{app:sec:exp_operator}
The operator representations are obtained by prompting GPT-4o with the prompt presented in Fig. \ref{fig:prompt_operator} with a temperature of 0 to reduce stochasticity. Prior to obtaining the operator representations, we perform co-reference resolution at a Chapter level resolution. Chapters are then chunked into smaller text chunks each containing three sentences without overlap between consecutive chunks. Space-Time coupling is performed after the operator representations are obtained by prompting GPT-4o with the prompt presented in Fig. \ref{fig:prompt_spacetime} with temperature set to 0 and max generation tokens set to 1000.

\subsection{Reconciler}
\label{app:sec:exp_reconciler}

Reconciliation is performed on consecutive chunks of operator representations; for our study, we reconcile all chunks of a particular chapter to produce one reconciled GSW representation per chapter. Roles and states for reconciled entities are time-stamped and stored, and this historical information is subsequently utilized during the generation of entity-level summaries.

When a reconciled entity provides new space/time information, its associated space/time nodes are updated accordingly. All previously recorded space/time information is also time-stamped and preserved to enrich these entity-level summaries. Furthermore, it is important to note that if an update to a space/time node is triggered by one entity, this new spatio-temporal information is propagated to all other entities coupled with that same node; this dynamic is illustrated in Figures \ref{app:fig:gsw_example2} and \ref{app:fig:gsw_example3}.

Finally, the reconciliation process also addresses \textit{forward-falling questions} —queries identified by previous Operator instances that can now be answered using the integrated information from the reconciled GSW as detailed in Section 2.1 of the main paper. These questions are resolved by prompting GPT-4o with the instructions detailed in Figure \ref{fig:prompt_reconcile}. For this QA resolution task, the temperature is set to 0 and maximum generation tokens are set to 500.

\subsection{QA}
\label{app:sec:exp_qa}

Prior to the final question answering (QA) stage, entity-specific summaries are generated using the GSW structure. For each entity, a prompt is constructed incorporating its roles, states, associated spatio-temporal information, and the questions it addresses through verb phrases (as captured in its GSW representation). This summarization prompt, detailed in Figure \ref{fig:prompt_summaries}, is processed by GPT-4o with a temperature of 0 and a maximum of 500 generation tokens.

The question answering (QA) process unfolds as follows: First, Named Entity Recognition (NER) is performed on the input question to identify relevant entities for querying the GSW. Based on these extracted entities, basic string matching is used to find corresponding entities within the consolidated GSW representations. Next, the \textbf{entity-specific summaries} (generated as described previously) for these matched entities are retrieved and then re-ranked. This re-ranking is based on the cosine similarity between the embeddings of the entity summaries and the embedding of the input query. To ensure consistency, the Voyage-03 model  is employed as the embedding model for both the summaries and the query. Finally, these re-ranked summaries are passed to the answering agent (GPT-4o ). The context provided to the agent is limited to summaries derived from a maximum of 17 diverse chapters, a constraint applied to maintain consistency across all evaluated methods  and to ensure all dataset questions can be addressed. A detailed example of the QA process is presented in Appendix \ref{app:sec:qa_example}.

\subsection{Baselines}
For HippoRAG2 \cite{JimenezGutierrez2025HippoRAG}, GraphRAG \cite{Edge2025GraphRAG}, and LightRAG \cite{Guo2024LightRAG}, we adhere to each method's default hyperparameters and prompt formats as provided in their respective official implementations. To ensure consistency across baselines, we modify the answering model in HippoRAG2 to use GPT-4o, aligning it with other evaluated methods. Additionally, we set top-k to 17 for HippoRAG2 to retrieve the top 17 relevant documents to align with the QA settings. The detailed configurations for each baseline are listed in Tables\ref{tab:graphrag-settings}--\ref{tab:lightrag-settings}.
\input{Tables/app_baseline_parameters}

\subsection{Bootstrapping for Evaluation}

In our main evaluation for EpBench-200 and EpBench-2000, we represent error bars computed via bootstrap resampling on 1,000 iterations. For each evaluation, we resampled the test set predictions with replacement and computed performance metrics on each bootstrap sample. The LLM judge operated with temperature=0 for deterministic outputs. These standard deviations indicate the variability in scores when different combinations of test examples are weighted through resampling

\section{Ablation Studies}
\label{app:sec:ablation}
We present the results of ablation studies we performed on our GSW framework.

\subsection{Evaluating the GSW on the Short Book Dataset}
Table \ref{Table:appendix_metric_grouped_short_book} presents results comparing GSW against Vanilla LLM on the shorter 20-chapter variant of EpBench. Both GSW and Vanilla LLM demonstrate strong performance on this smaller dataset. The Vanilla LLM performs particularly well on this version because the entire context length is approximately 10,000 tokens, which easily fits within the model's context window. Notably, even with this shorter context, we observe that Vanilla LLM begins to struggle relative to GSW as the number of matching cues increases, particularly in the 3-5 cue category where GSW shows superior recall (0.910 vs 0.781) and F1-score (0.857 vs 0.777).

This finding further supports our main results presented in Table 2 of the main paper, as it demonstrates how Vanilla LLM's performance deteriorates with increased context length. While performing competitively on short narratives, Vanilla LLM struggles with the 200-chapter version where context exceeds 100,000 tokens. In contrast, GSW maintains robust performance across both short and long narratives , highlighting the value of our approach.
\input{Tables/app_Epbench_20}

\subsection{Detailed Results on EpBench-2000}

The detailed statistics for EpBench-2000 are presented in Table \ref{tab:dataset_stats_2000}. Although the maximum number of chapters referenced per query in the EpBench-2000 dataset reaches 138, we choose to limit the maximum context utilization to 17 chapters per query, maintaining the same configuration applied to EpBench-200 in the main paper. This choice is based on the fact that the 138-chapter scenario represents an extreme outlier, while 17 chapters suffice to address the majority of queries effectively. Furthermore, processing 138 chapters per query would introduce significant computational overhead and inefficiencies, as it requires feeding an excessive volume of text to the model, which could negatively impact both performance and resource utilization. Since we use the same number of chapters per query as in EpBench-200, we therefore expect a very similar token usage.

Table \ref{Table:epbench_2000_bootstrap} reports the complete set of metrics for GSW and all baselines on the EpBench-2000 dataset, broken down by cue complexity. These results expand upon the summary in the main text, demonstrating that GSW retains its lead across all levels of episodic complexity, and outperforming the strongest baseline by more than \textbf{15\%} in F1-score and \textbf{14\%} in recall. The EpBench-2000 experiment further highlights GSW’s ability to scale effectively while maintaining strong performance in long-context, high-recall settings.
\input{Tables/data_content_2000}
\input{Tables/epbench_2000_bootstrap}
\subsection{Ablating components of the GSW for Question Answering}

\input{Tables/app_ablation_gsw}

Table ~\ref{Table:gsw_ablation_study_fully_filled} presents the results of ablating both components of the GSW as well as approaches to retrieval, highlighting the importance of each component and our string matching + reranking retrieval mechanism. We note that while naive string matching achieves almost similar performance to our retrieval method, it consumes almost double the number of tokens.

\section{Related work on Memory Augmentation for LLMs}
\label{app:sec:relatedmemory}

Enabling LLMs to effectively process long narratives requires capabilities akin to human episodic memory – constructing and maintaining a dynamic, coherent understanding of events unfolding over space and time \cite{tulving_episodic_1972, eichenbaum_corticalhippocampal_2000}. Key to this is the ability to accurately track entities, including their evolving states and roles, and to ground events and answer queries based on specific spatial and temporal contexts established within the narrative \cite{huet_episodic_2025}. While LLMs possess remarkable core abilities, achieving this level of sophisticated, stateful reasoning over extended sequences remains a significant challenge. The following sections analyze inherent limitations in common approaches used to provide context to LLMs, evaluating why they often fall short of systematically delivering these specific episodic memory capabilities.

\subsection{Leveraging Long context LLMs}

One approach to providing LLMs with relevant context is to leverage their increasingly large context windows, potentially feeding the entire long narrative along with a query into the prompt. The rapid expansion of context lengths, now reaching millions of tokens, has certainly broadened the scope of tasks LLMs can handle by allowing more raw information to be processed simultaneously \cite{team2024gemini}.

However, relying solely on this native processing mechanism faces significant hurdles when evaluated against the demands of episodic memory. Firstly, while context windows are growing, they are not infinite, and extremely long narratives may still exceed even the largest available limits. Secondly, even when a narrative technically fits, processing vast amounts of text for every query is computationally expensive, impacting latency and cost. More fundamentally, processing quality often degrades with extreme context lengths \cite{leng2024long, hsieh_ruler_2024, wang2024multimodal}. Research has shown that LLMs can struggle to consistently access and utilize information spread across very long contexts, with performance notably dipping for information located in the middle (\textit{lost in the middle} phenomenon) \cite{liu_lost_2023}. Feeding potentially large amounts of irrelevant text alongside the crucial details for a specific episodic query can distract the model and hinder its ability to pinpoint and reason over the necessary information.

Finally, perhaps the most critical limitation for systematic episodic tracking is the inherently unstructured nature of the input context. Even with all the necessary details about entity states, roles, locations, and times present within the text, the LLM lacks explicit mechanisms to structure this information dynamically. It must rely solely on its attention mechanism and in-context learning to piece together relationships, track state changes, and maintain temporal coherence across potentially thousands of tokens. This makes the reliable, systematic tracking required for robust episodic memory challenging and often brittle when relying only on the native context window \cite{huet_episodic_2025}.

\subsection{Memory Augmentation for LLMs}

To overcome the challenges of static parametric knowledge and the inefficiencies of processing entire documents in context, Retrieval-Augmented Generation (RAG) has become a standard technique \cite{lewis_retrieval-augmented_2021, gao_retrieval-augmented_2024}. The typical RAG pipeline involves pre-processing a knowledge corpus (e.g., the entire narrative document) into smaller chunks. These chunks are then indexed, commonly using dense vector embeddings obtained from encoder style LLMs\cite{bert, reimers_sentence-bert_2019, lee_nv-embed_2025}, though sparse methods like BM25\cite{robertson_probabilistic_2009} or hybrid approaches are also employed \cite{cormack2009reciprocal}. At inference time, the user query is used to retrieve the top-k most relevant chunks from the index based on a similarity metric (e.g., cosine similarity for dense vectors). These retrieved chunks are then presented as augmented context to an LLM, which generates the final response based on both its parametric knowledge and the retrieved information.\cite{ram_-context_2023}

This approach has proven effective for many knowledge-intensive tasks, particularly fact-based question answering where retrieving specific evidence snippets is sufficient \cite{karpukhin-etal-2020-dense}. However, when evaluated against the requirements of robust episodic memory recall over long narratives, the limitations of standard RAG become apparent \cite{huet_episodic_2025}. Firstly, the process of retrieving discrete, potentially disconnected chunks based on local query relevance often \textbf{fragments the narrative flow}. This makes it exceedingly difficult for the LLM to reliably follow evolving storylines or track the \textbf{changing states and roles of entities} over time, as the necessary context may be spread across multiple chunks that are not retrieved together\cite{chen_walking_2023}.

Moreover, this fragmentation problem is compounded by the framework being highly sensitive to the initial chunking strategy\cite{merola2025reconstructing}. Arbitrary chunk boundaries can split crucial information related to an event or an entity's state, leading to information loss during retrieval. For instance, if a character's state changes within a passage, but the chunking algorithm divides this passage at an inopportune point, the complete context of this state change may not be captured in any single retrieved chunk. Optimal chunking is non-trivial and can significantly impact the ability to reconstruct the necessary context for complex episodic reasoning. Consequently, while standard RAG offers efficiency gains over naive long-context processing, its inherent lack of structure and narrative coherence makes it ill-suited for systematically addressing the dynamic, stateful, and context-dependent nature of episodic memory tasks.

Additionally, standard RAG mechanisms based on semantic similarity often struggle with incorporating specific spatio-temporal constraints that are essential for episodic memory. Embeddings typically capture semantic content but may not adequately encode the nuances of time and location, making it difficult to retrieve context relevant to a specific point in time or place mentioned in a query or implied by the narrative history.

\subsection{Structured Representations as Memory}

Recognizing the limitations of standard RAG, particularly its tendency to fragment narratives and struggle with temporal coherence, recent work has explored incorporating more explicit structure into the retrieval and augmentation process. Instead of treating the source narrative as a flat sequence of independent chunks, these methods attempt to build richer representations that capture relationships or hierarchies within the text, aiming to provide more contextually relevant information to the LLM.

While these structured approaches offer advantages over standard RAG by preserving more relational or hierarchical context and enabling more sophisticated information integration (like multi-hop reasoning or global summarization), they still face challenges when viewed through the lens of episodic memory \cite{huet_episodic_2025}. Graph-based methods like GraphRAG\cite{Edge2025GraphRAG}, LightRAG\cite{Guo2024LightRAG}, HippoRAG\cite{JimenezGutierrez2025HippoRAG, gutierrez_hipporag_2025} and RAPTOR\cite{sarthi_raptor_2024} suffer from two broad limitations. First, they lack mechanisms to track entity state/role changes across time—they represent entities as static nodes without modeling how attributes or relationships evolve throughout a narrative. Second, they provide no specific framework to ground the evolving narrative in space and time, making it difficult to represent sequential developments or causal relationships. These methods typically represent semantic relationships or summarize community structures within a potentially static corpus, but they are not explicitly designed to model the temporal flow of events within a single narrative or to meticulously track the dynamic changes in entity states and roles as the narrative unfolds sequentially. Their structure captures connections, but not necessarily the chronological progression and state transitions required for recalling specific episodes.

Other research efforts have targeted episodic memory more directly. For instance, Larimar \cite{das_larimar_2024} proposes modifications to the LLM's attention mechanism, while EM-LLM \cite{fountashuman} introduces specific memory components integrated with openweight models. While promising, these approaches often require fundamental changes to the LLM architecture or are designed specifically for openweight models, limiting their applicability. In contrast, our GSW framework is proposed as a plug-and-play episodic memory module compatible with any underlying LLM (including closed-source models like GPT-4o via API) and, critically, requires no specialized training or fine-tuning of model parameters, relying instead on the LLM's capabilities for its operator and reconciliation functions.

\section{Computational Costs and Resources for Building the GSW}
\label{app:sec:CompCost}

The primary computational costs for the Generative Semantic Workspace (GSW) framework are associated with its initial, one-time indexing process. To index the 200 chapters of the Epbench dataset, the total expense is approximately \$15 when utilizing GPT-4o. This cost covers all stages of GSW construction, including the generation of operator representations, reconciliation, and the creation of entity-specific summaries. By leveraging parallel calls to the OpenAI API, managed via the Bespoke Curator library \cite{bespoke_curator}, this entire indexing task for 200 chapters can be completed in roughly 1 hour. Alternatively, the OpenAI Batch API can be used to reduce costs, with indexing taking hours. 

Our primary experiments leverage API-based models (e.g., GPT-4o) and therefore do not necessitate dedicated local computing infrastructure. However, for tasks such as running the baseline method evaluations reported in this study, and for broader experimentation involving various dense retriever models or locally-hosted chat models, we utilized a single server node equipped with four A6000 GPUs.

\section{Related Computational Models of Semantics}
\label{app:sec:relatedmodels}
Semantic representation frameworks have a rich history in NLP, yet as we explore below, their design choices create inherent limitations for tracking evolving actor states and relationships—a critical requirement for episodic memory. Among the most influential frameworks are PropBank~\cite{propbank} and FrameNet~\cite{framenet}, which attempt to define correspondences between (a) the syntactic ``realizations'' of semantics \textit{explicit} within language structure, and (b) finite, discrete sets of semantic ``roles''~\cite{levin}. These approaches rely heavily on manually-annotated lexicon ontologies developed by expert linguists. While valuable for understanding individual sentences, they were not designed for the dynamic, interconnected tracking that episodic memory demands. Below, we detail these frameworks and their limitations for serving as memory systems:

\noindent \textbf{PropBank:} PropBank utilized a \textit{bottom-up} approach: (1) Dependency Parse Trees~\cite{dpt} were applied to a large text corpus to distill shared syntactic patterns (``Framesets'') specific to each verb (a process known as ``lexical sampling''). (2) For each Frameset, the corresponding sentences were manually annotated with an enumerated set of \textit{arguments} \texttt{ARG:0,\dots, ARG:N}. These arguments were later associated to verb-specific definitions using VerbNet~\cite{verbnet}. The  semantic roles are identified as corresponding \textit{spans} within the sentence (commonly a \texttt{NP}, \texttt{NNP} subtree in the dependency parsing). For example, the sentence (A):
    \begin{center}

    \noindent \textbf{Officers} captured \textbf{Sarah} at the \\ \textbf{Sepulveda on-ramp} of the \textbf{405}.
    \end{center}
would be annotated with the arguments: 
\begin{center}

\noindent \textit{Agent}: \textbf{officers}, \textit{Predicate}: \textbf{captured}, \textit{Patient}: \textbf{Sarah}.
\end{center}
Perhaps the greatest benefit of PropBank was that its syntactic ``grounding'' made it possible for rule-based and early ML models~\cite{rml, srlbert} to \textit{learn the task of distilling the semantics} given a sentence, albeit within the confines of a \textit{ limited} ontology of $>3000$ verbs and $>4000$ Framesets 

\noindent \textit{Event Databases:} PropBank evolved in several directions, including efforts to unify it with related semantic lexicon such as VerbNet and FrameNet~\cite{semlink, semlink2}, or augment it via the DWD overlay~\cite{dwdoverlay} to WikiData~\cite{wikidata}. The latter of these efforts now manifests as ``Event'' databases~\cite{eventreview, xiang2019survey} such as the ACE~\cite{ace} and ERE~\cite{ere} datasets, and led to the DARPA initiative of Event identification/extraction challenges. Events are best motivated by their related identification tasks: Given a sentence, identify the event(s) -- from a set of \textit{hundreds} of events in a pre-annotated schema~\cite{glen, eventextract, text2event} -- that the sentence is referring to. For example, (A) would be annotated with the \textit{Capture} event.

\noindent \textbf{FrameNet:} In contrast to PropBank and related Event ontologies, FrameNet\footnote{\url{https://framenet.icsi.berkeley.edu/}} utilizes a \textit{top-down} approach that is not tethered to the syntax structure. Rather, expert linguists aggregated roles (redefined as ``Frame Elements'' (FE)) from a large corpus of sentences, which are \textit{known to co-exist} under a conceptual gestalt, or ``Frame''. Each frame additionally comprises a set of ``Lexical Units'' (LU) - valences (mostly verbs and nouns) whose occurrence in a sentence increases the likelihood of a frame. For example,
\begin{center}
the \textit{Frame}: \textbf{Taking Captive} 
\end{center}
would contain the following frame elements and lexical units: 
\begin{center}
FE: \textit{Agent}, \textit{Captive}, \textit{Cause} \\
LU: \textit{capture.v}, \textit{secure.v}
\end{center}

 FrameNet (thousands of frames and tens of thousands of FE) is a substantially larger and more comprehensive ontology~\cite{framenetcompare} compared to Propbank. When originally constructed, automated systems could not effectively identify the frames implied by a sentence; today, however, Transformer models~\cite{attention, bert, fst} have demonstrated success at accurately modeling the sentence-to-frame mapping.

Despite the enormous success and wide adoption of PropBank, FrameNet, and their descendant works, the explicit, finite, and discrete lexicons they employ raise the question: \textit{When is an explicit lexicon ontology complete?}.  While FrameNet provides Frame-Frame precedence and subset relationships, these are coarse-grained and do not adequately answer the question: \textit{How can we track the evolution of semantics across a stream of sentences?} - a key requirement for any semantic model to serve as a memory.

More recent work ~\cite{conceptnet} has attempted to assemble a comparatively larger (and less stringent) open-schema semantic ontology of concepts using game-play based crowd-sourcing techniques~\cite{verbosity}. However, such efforts to scale manual annotation ultimately do not address how a complete ontology can be constructed. Event Graph Models (EGM)~\cite{tegm} generate event networks to describe the dynamics of events in a text corpus, often using a combination of submodules such as Coreference Resolution~\cite{coref}, Named Entity Recognition (NER)~\cite{ner} and Semantic Role Labeling~\cite{srl}. Extensions~\cite{future} generate the \textit{most likely} event \textit{template} sequences. These methods rely on predefined event schema to enumerate the set of possible events. However, while EGMs both track the evolution of semantics across sentences and offer an unsupervised approach to extending existing ontologies, these often marginalize across individual contexts in the training corpus, and generate the most likely event \textit{schema} that follows a current event schema network. As a result, these works have yet to design methods to track the semantics across a specific document. The GSW, particularly through the operator, is designed to overcome these challenges by generating actor-centric, evolving semantic maps that are not constrained by predefined, static lexicons and can capture the nuances of unfolding situations.To demonstrate the GSW Operator's effectiveness in producing these comprehensive semantic maps from complex, real-world text, we conducted a qualitative human evaluation.

\subsection{Comparing Existing Semantic Models to the Operator}

To empirically validate the Operator's capability in generating these comprehensive semantic maps, particularly its proficiency in interpreting narrative-rich texts where actor, roles and states undergo significant evolution, we leveraged news reports, as they are a popular resource for sampling semantics-rich stories that belong to universally recognized situation patterns.
We query GDELT~\cite{gdelt}, a Jigsaw-powered news-indexing platform, with Situation identifiers to retrieve a small set of situation-conditioned \texttt{en\_US} articles. Table ~\ref{tab:operator_data} presents statistics about the data. These situations were manually selected as an initial seed set -- similar to FrameNet's early versions containing few frames~\cite{fillmore} -- to assess the validity of the GSW framework. Situation-specific seeds are assembled using a bootstrapped method that invokes FrameNet: (a) Frames are linked using subframe and precedence relationships to create weakly connected components; (b) Headers/labels of the frames in each component form the seed search phrases. We evaluate our framework on situations like ``Crime and Justice'', ``Fire Fighting'', ``Healthcare'', and ``Technology Development''.

\input{Tables/operator_dataset_stats}

Table \ref{tab:operator-framework} presents these results across five diverse situations, showing strong human preference for the Operator generated representations compared to existing semantic frameworks.

\input{Tables/operatorframeworks}

\subsection{Annotator Guidelines}

\begin{figure*}[t]
\centering
\includegraphics[width=0.8\textwidth]{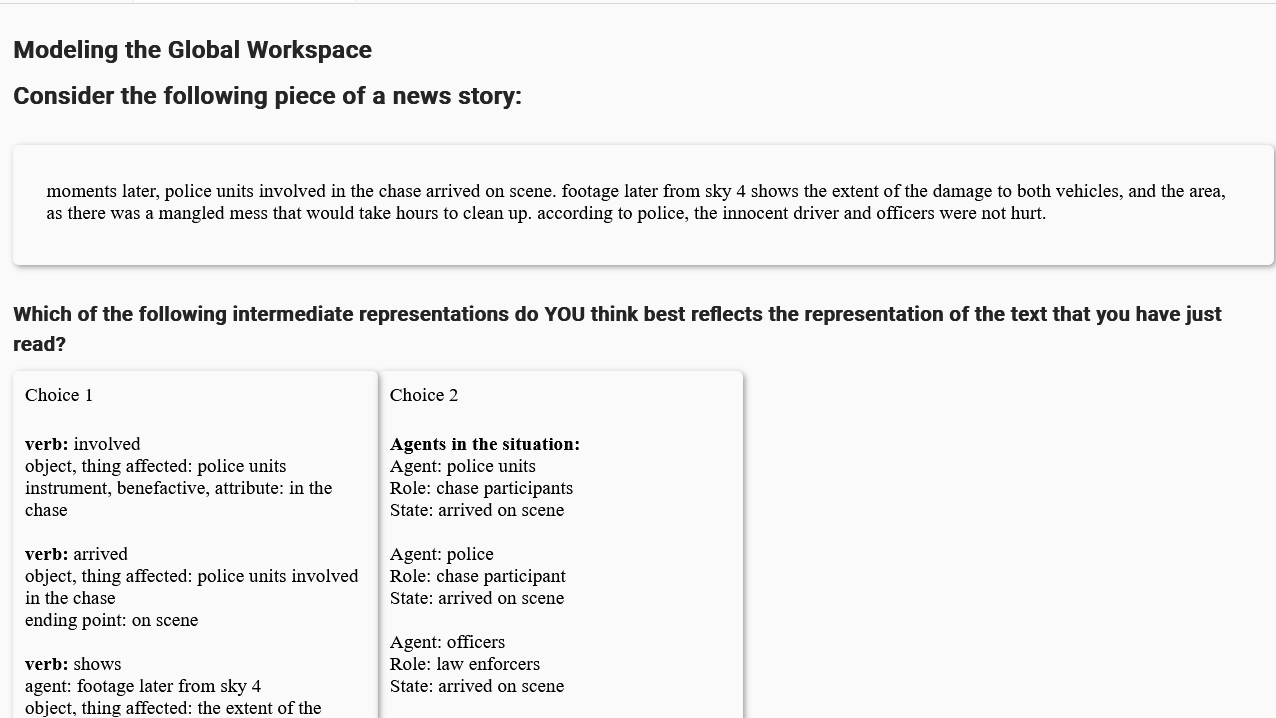}
\caption{\textbf{Annotator instructions for UpWork Task:} Annotators are asked to compare the outputs of the Operator to the Semantic map output by a baseline framework (either GLEN, BertSRL, FST) given a shared input text context. During annotation, one random baseline map and the Operator output are presented in random order and the annotator is asked to pick the representation of the Semantics that best reflects the information in the context.}
\label{fig:ann}
\end{figure*}

Annotators who exceeded $\$50K$ in total gross pay were recruited from UpWork, a talent resource. These candidates were first interviewed in a $10$-minute session to verify that they were proficient in English and those that had prior experience in annotating large-scale AI/ML data -- listed as a verified skill on the platform -- were selected to move on to the next round. Following this, they were given a set of $10$ task prototype examples and $10$ unanswered labeling tasks. Those that got $9$ out of the $10$ annotations right moved on to the first round of labeling. Each task was labeled twice -- by the \textit{annotator} and a \textit{verifier} -- to ensure quality of the results. Annotators were paid $\$5 / 40 \textrm{ samples}$ which was estimated to take them about $30$ minutes at most, or at the rate of $\$10 / \textrm{hour}$, which was confirmed to exceed the federal minimum wage where the annotators were situated. Annotator guidelines are presented in Fig.~\ref{fig:ann}

\clearpage

%% file: Tables/app_sectionD_Queries.tex
\begin{table*}[!htbp] 
  \centering
  \scriptsize 
  \begin{tabularx}{\linewidth}{@{}l >{\raggedright\arraybackslash}X >{\raggedright\arraybackslash}X@{}}
    \toprule
    \textbf{Query ID} & \textbf{Query Text} & \textbf{Ground Truth Answer} \\
    \midrule
    Q1 & Consider all events that Jackson Ramos has been involved in. List all the locations where these events took place, without mentioning the events themselves. & High Line, Snug Harbor Cultural Center, Central Park, One World Trade Center, Ellis Island \\
    \addlinespace 
    Q2 &  Reflect on the experiences of Carter Stewart related to Scientific Conference. List all the unique locations where these events took place, without mentioning the events themselves. & Bethpage Black Course, Metropolitan Museum of Art \\
    \addlinespace
    Q3 &Consider all events that Ezra Edwards has been involved in. List all the locations where these events took place, without mentioning the events themselves. & Water Mill Museum, Port Jefferson, Yankee Stadium, New York Botanical Garden, Brooklyn Bridge, Bethpage Black Course, One World Trade Center \\
    \addlinespace
    Q4 & Recall the events related to Tech Hackathon that occurred on March 23, 2025. List all the locations where these events took place, without describing the events themselves. & Yankee Stadium, Water Mill Museum, Woolworth Building, Queensboro Bridge \\
    \addlinespace
    Q5 & Recall the events related to Tech Hackathon that occurred on November 13, 2026. List all the locations where these events took place, without describing the events themselves. & Trinity Church, Woolworth Building, Statue of Liberty, Fire Island National Seashore \\
    \bottomrule
  \end{tabularx}
  \caption{Selected Queries and Ground Truth Answers for Qualitative Analysis}
  \label{tab:qual_queries_answers}
\end{table*}

%% file: Tables/app_sectionD_perf.tex
\begin{table*}[!htbp]
  \centering
  \scriptsize 
  \begin{tabularx}{\linewidth}{@{}llc >{\raggedright\arraybackslash}X >{\raggedright\arraybackslash}X@{}}
    \toprule
    \textbf{Query ID} & \textbf{Method} & \textbf{Token Count} & \textbf{Error Description} & \textbf{Analysis/Reason} \\
    \midrule
    \multirow{3}{*}{Q1} & GSW           & 2011 & None & NA \\
                        & HippoRAG2     & 9289 & None & NA \\
                        & GraphRAG      & 8189 & Missing 1 location & Info not available in retrieved context. \\
    \midrule 
    \multirow{3}{*}{Q2} & GSW           & 1568 & None & NA \\
                        & HippoRAG2     & 8225 & Hallucinated 3 extra locations. & Too much irrelevant information resulted in LLM hallucination. \\
                        & GraphRAG      & 8220 & Missed 1 location and Hallucinated 2 & All required info present in context but LLM hallucinated. \\
    \midrule 
    \multirow{3}{*}{Q3} & GSW           & 1726 & None & NA \\
                        & HippoRAG2     & 8475 & Missed 1 location & Info not available in retrieved context. \\
                        & GraphRAG      & 7058 & Missed 1 location & Info not available in retrieved context. \\
    \midrule 
    \multirow{3}{*}{Q4} & GSW           & 5530 & None & NA \\
                        & HippoRAG2     & 8614 & Missed 2 locations & All required info present in context but LLM hallucinated. \\
                        & GraphRAG      & 7936 & Missed 3 locations and Hallucinated 1 & All required info present in context but LLM hallucinated. \\
    \midrule 
    \multirow{3}{*}{Q5} & GSW           & 6452 & None & NA \\
                        & HippoRAG2     & 8355 & Missed 1 location & Info not available in retrieved context \\ 
                        & GraphRAG      & 7936 & Missed 2 locations & Info not available in retrieved context. \\
    \bottomrule
  \end{tabularx}
  \caption{Qualitative Performance Comparison on Selected Queries (referencing Query IDs from Table \ref{tab:qual_queries_answers})}
  \label{tab:qual_analysis_comparison}
\end{table*}

%% file: Tables/app_baseline_parameters.tex
\begin{table}[ht]
\centering
\scriptsize
\begin{tabular}{@{}ll@{}}
\toprule
\textbf{Setting} & \textbf{Value} \\
\midrule
Mode & Local \\
LLM Model & \texttt{gpt-4o} \\
Embedding Model & \texttt{text-embedding-3-small} \\
Response Type & Multiple paragraphs \\
Max Context Tokens & 12000 \\
Text Unit Proportion & 0.5 \\
Community Report Proportion & 0.1 \\
Top-K Entities & 10 \\
Top-K Relationships & 10 \\
Include Entity Rank & True \\
Include Relationship Weight & True \\
Include Community Rank & False \\
\bottomrule
\end{tabular}
\caption{GraphRAG Baseline Parameter}
\label{tab:graphrag-settings}
\end{table}

\begin{table}[ht]
\centering
\scriptsize
\begin{tabular}{@{}ll@{}}
\toprule
\textbf{Setting} & \textbf{Value} \\
\midrule
LLM Indexing Model & \texttt{gpt-4o-mini} \\
LLM Answering Model & \texttt{gpt-4o} \\
Embedding Model & \texttt{NV-Embed-v2} \\
QA Top-K & 17 \\
Linking Top-K & 5 \\
Retrieval Top-K & 200 \\
\bottomrule
\end{tabular}
\caption{HippoRAG2 Baseline Parameter}
\label{tab:hipporag-settings}
\end{table}

\begin{table}[ht]
\centering
\scriptsize
\begin{tabular}{@{}ll@{}}
\toprule
\textbf{Setting} & \textbf{Value} \\
\midrule
LLM Model & \texttt{gpt-4o} \\
Embedding Model & \makecell[l]{\texttt{text-embedding-3-}\\\texttt{small}} \\
Retrieval Mode & Hybrid \\
Chunk Token Size & 1200 \\
Chunk Overlap Size & 100 \\
\bottomrule
\end{tabular}
\caption{LightRAG Baseline Parameter}
\label{tab:lightrag-settings}
\end{table}

%% file: Tables/app_Epbench_20.tex
\begin{table*}[t] 
\centering
\small 
\setlength{\tabcolsep}{4pt} 
\begin{tabular}{@{}llccccc@{}}
\toprule
\multirow{2}{*}{Metric} & \multirow{2}{*}{Method} & 0 Cues & 1 Cue & 2 Cues & 3-5 Cues & Overall \\
& & (N=180) & (N=180) & (N=72) & (N=24) & (N=456) \\
\midrule
\multirow{2}{*}{\textbf{P}} & Vanilla LLM & 0.889 & \textbf{0.781} & \textbf{0.900} & 0.799 & \textbf{0.843} \\
& GSW (Ours) & \textbf{0.939} & 0.751 & 0.804 & \textbf{0.854} & 0.841 \\
\midrule 
\multirow{2}{*}{\textbf{R}} & Vanilla LLM & 0.889 & \textbf{0.919} & 0.813 & 0.781 & 0.883 \\
& GSW (Ours) & \textbf{0.939} & 0.856 & \textbf{0.819} & \textbf{0.910} & \textbf{0.886} \\
\midrule 
\multirow{2}{*}{\textbf{F1}} & Vanilla LLM & 0.889 & \textbf{0.812} & \textbf{0.821} & 0.777 & \textbf{0.842} \\
& GSW (Ours) & \textbf{0.939} & 0.745 & 0.784 & \textbf{0.857} & 0.834 \\
\bottomrule
\end{tabular}
\caption{\textbf{Full EpBench (20-Chapter Book) Performance by Event Categories:} Precision (P), Recall (R), and F1-Score for Vanilla LLM vs. GSW across different event category complexities. (N=X) indicates questions per category.}
\label{Table:appendix_metric_grouped_short_book}
\end{table*}

%% file: Tables/data_content_2000.tex
\begin{table}[htbp]
\centering
\small
\begin{tabular}{@{}ll@{}} 
\toprule
Statistic                               & Value \\
\midrule
Number of Chapters                      & 1967 \\
Total Tokens                            & 1,012,097 \\
Total Queries (QA Pairs)                & 623 \\
Queries by Event Category               & \\ 
\quad (0 / 1 / 2 / 3-5 / 6+ Cues)     & 90 / 165 / 114 / 124 / 130 \\
\midrule
Max. Chapters Referenced per Query      & 138 \\
Min. Chapters Referenced per Query      & 0 \\
\bottomrule
\end{tabular}
\caption{\textbf{EpBench-2000 Dataset Statistics.}}
\label{tab:dataset_stats_2000}
\end{table}

%% file: Tables/epbench_2000_bootstrap.tex
\begin{table*}[t]
\centering
\scriptsize 
\setlength{\tabcolsep}{4pt} 
\begin{tabular}{@{}llcccccc@{}}
\toprule
\multirow{2}{*}{Metric} & \multirow{2}{*}{Method} & 0 Cues & 1 Cue & 2 Cues & 3-5 Cues & 6+ Cues & Overall \\
& & (N=90) & (N=165) & (N=114) & (N=124) & (N=130) & (N=623) \\
\midrule
\multirow{6}{*}{\textbf{P}}
& Embedding RAG & $0.789\pm0.043$ & $\underline{0.751}\pm0.028$ & $\mathbf{0.845}\pm0.026$ & $\underline{0.840}\pm0.031$ & $\mathbf{0.911}\pm0.025$ & $\underline{0.827}\pm0.014$ \\
& GraphRAG \cite{Edge2025GraphRAG} & $\mathbf{0.943}\pm0.025$ & $0.747\pm0.038$ & $0.639\pm0.040$ & $0.692\pm0.038$ & $0.795\pm0.043$ & $0.761\pm0.017$ \\
& HippoRAG2 \cite{JimenezGutierrez2025HippoRAG}  & $0.620\pm0.051$ & $0.638\pm0.032$ & $0.803\pm0.032$ & $0.824\pm0.028$ & $\underline{0.893}\pm0.021$ & $0.759\pm0.016$ \\
& LightRAG \cite{Guo2024LightRAG} & $0.790\pm0.042$ & $0.534\pm0.039$ & $0.560\pm0.040$ & $0.593\pm0.035$ & $0.787\pm0.039$ & $0.649\pm0.018$ \\
\cmidrule(lr){2-8}
& GSW (Ours) & $\underline{0.867}\pm0.0025$ & $\mathbf{0.761}\pm0.0020$ & $\underline{0.841}\pm0.0019$ & $\mathbf{0.841}\pm0.0019$ & $0.870\pm0.0019$ & $\mathbf{0.830}\pm0.0010$ \\

\midrule
\multirow{6}{*}{\textbf{R}} 
& Embedding RAG  & $0.789 \pm 0.043$ & $\underline{0.764} \pm 0.032$ & $\underline{0.795} \pm 0.033$ & $0.637 \pm 0.031$ & $0.480 \pm 0.028$ & $\underline{0.688} \pm 0.015$ \\
& GraphRAG \cite{Edge2025GraphRAG} & $\mathbf{0.943} \pm 0.025$ & $0.492 \pm 0.037$ & $0.587 \pm 0.039$ & $0.538 \pm 0.036$ & $0.321 \pm 0.025$ & $0.548 \pm 0.017$ \\
& HippoRAG2 \cite{JimenezGutierrez2025HippoRAG} & $0.620 \pm 0.050$ & $0.703 \pm 0.034$ & $0.769 \pm 0.031$ & $\underline{0.647} \pm 0.029$ & $\underline{0.491} \pm 0.026$ & $0.648 \pm 0.016$ \\
& LightRAG \cite{Guo2024LightRAG} & $0.790 \pm 0.042$ & $0.525 \pm 0.038$ & $0.549 \pm 0.038$ & $0.440 \pm 0.033$ & $0.270 \pm 0.017$ & $0.497 \pm 0.017$ \\
\cmidrule(lr){2-8}
& GSW (Ours) & $\underline{0.867} \pm 0.025$ & $\mathbf{0.844} \pm 0.019$ & $\mathbf{0.864} \pm 0.016$ & $\mathbf{0.792} \pm 0.017$ & $\mathbf{0.633} \pm 0.017$ & $\mathbf{0.796} \pm 0.009$ \\

\midrule
\multirow{6}{*}{\textbf{F1}} 
& Embedding RAG  & $0.789 \pm 0.043$ & $\underline{0.644} \pm 0.031$ & $\underline{0.758} \pm 0.032$ & $0.679 \pm 0.031$ & $0.561 \pm 0.029$ & $\underline{0.675} \pm 0.015$ \\
& GraphRAG \cite{Edge2025GraphRAG} & $\mathbf{0.943} \pm 0.025$ & $0.436 \pm 0.035$ & $0.547 \pm 0.038$ & $0.541 \pm 0.036$ & $0.405 \pm 0.027$ & $0.544 \pm 0.017$ \\
& HippoRAG2 \cite{JimenezGutierrez2025HippoRAG} & $0.620 \pm 0.050$ & $0.583 \pm 0.031$ & $0.732 \pm 0.031$ & $\underline{0.681} \pm 0.027$ & $\underline{0.578} \pm 0.026$ & $0.635 \pm 0.015$ \\
& LightRAG \cite{Guo2024LightRAG} & $0.790 \pm 0.042$ & $0.436 \pm 0.034$ & $0.514 \pm 0.037$ & $0.463 \pm 0.033$ & $0.375 \pm 0.021$ & $0.494 \pm 0.016$ \\
\cmidrule(lr){2-8}
& GSW (Ours) & $\underline{0.867} \pm 0.025$ & $\mathbf{0.741} \pm 0.020$ & $\mathbf{0.818} \pm 0.017$ & $\mathbf{0.789} \pm 0.017$ & $\mathbf{0.698} \pm 0.016$ & $\mathbf{0.773} \pm 0.009$ \\

\bottomrule
\end{tabular}
\caption{\textbf{GSW performance on Epbench (2000-Chatpers Book):} Performance is grouped by metric (Precision, Recall, F1-Score) across different numbers of matching cues per query. (N=X) indicates questions per category. Error bars are estimated via bootstrap resampling. Best score in each column for each metric group is \textbf{bold}; second best is \underline{underlined}.}

\label{Table:epbench_2000_bootstrap}
\end{table*}

%% file: Tables/app_ablation_gsw.tex
\begin{table*}[t] 
\centering
\small
\setlength{\tabcolsep}{3pt} 
\begin{tabular}{@{}llcccccc@{}}
\toprule
\multirow{2}{*}{Metric} & \multirow{2}{*}{GSW Configuration / Ablation} & 0 Events & 1 Event & 2 Events & 3-5 Events & 6+ Events & Overall \\
& & (N=150) & (N=150) & (N=90) & (N=98) & (N=60) & (N=548) \\
\midrule
\multirow{5}{*}{\textbf{P}} 
& w/o Space/Time Linking & 0.978 & 0.799 & 0.814 & 0.851 & 0.854 & 0.868 \\
& QA Input: Verb Phrases & 0.939 & 0.839 & 0.807 & 0.896 & 0.874 & 0.878 \\
& Retrieval: Str.Match, No reranking & 0.967 & 0.773 & 0.860 & 0.872 & 0.932 & 0.879 \\ 
& Retrieval: Emb. Match, No reranking & 0.922 & 0.792 & 0.797 & 0.874 & 0.876 & 0.855 \\ 
& Retrieval: NER emb, no reranking & 0.944 & 0.747 & 0.823 & 0.872 & 0.854 & 0.854 \\
\cmidrule(lr){2-8} 
& \textbf{GSW (Full)} & \textbf{0.978} & \textbf{0.755} & \textbf{0.810} & \textbf{0.878} & \textbf{0.891} & \textbf{0.865} \\
\midrule 
\multirow{5}{*}{\textbf{R}} 
& w/o Space/Time Linking & 0.978 & 0.800 & 0.810 & 0.738 & 0.723 & 0.827 \\
& QA Input: Verb Phrases & 0.939 & 0.766 & 0.644 & 0.674 & 0.551 & 0.747 \\
& Retrieval: Str.Match, No reranking & 0.967 & 0.834 & 0.850 & 0.819 & 0.822 & 0.867 \\
& Retrieval: Emb. Match, No reranking & 0.922 & 0.820 & 0.833 & 0.825 & 0.781 & 0.845 \\
& Retrieval: NER emb, no reranking & 0.944 & 0.710 & 0.750 & 0.721 & 0.624 & 0.768 \\
\cmidrule(lr){2-8} 
& \textbf{GSW (Full)} & \textbf{0.978} & \textbf{0.863} & \textbf{0.868} & \textbf{0.892} & \textbf{0.822} & \textbf{0.894} \\
\midrule 
\multirow{5}{*}{\textbf{F1}} 
& w/o Space/Time Linking & 0.978 & 0.731 & 0.764 & 0.762 & 0.761 & 0.811 \\
& QA Input: Verb Phrases & 0.939 & 0.733 & 0.633 & 0.719 & 0.621 & 0.754 \\
& Retrieval: Str.Match, No reranking & 0.967 & 0.748 & 0.826 & 0.823 & 0.859 & 0.846 \\
& Retrieval: Emb. Match, No reranking & 0.922 & 0.726 & 0.788 & 0.827 & 0.810 & 0.817 \\
& Retrieval: NER emb, No reranking & 0.944 & 0.629 & 0.717 & 0.748 & 0.693 & 0.756 \\
\cmidrule(lr){2-8} 
& \textbf{GSW (Full)} & \textbf{0.978} & \textbf{0.745} & \textbf{0.806} & \textbf{0.867} & \textbf{0.834} & \textbf{0.850} \\
\bottomrule
\end{tabular}
\caption{\textbf{Ablation Study of GSW Components on EpBench (200-Chapter Book):} Performance across different event categories (Precision, Recall, F1-Score). (N=X) indicates questions per category. Full GSW model results (at the bottom) are for reference from Table 2 in the main paper.}
\label{Table:gsw_ablation_study_fully_filled}
\end{table*}

%% file: Tables/operator_dataset_stats.tex
\begin{table}[h]
\centering
\small\addtolength{\tabcolsep}{-3pt}
\begingroup
\renewcommand{\arraystretch}{0.4} 
\begin{tabular}{@{}c|c|c|c@{}}
\toprule
\textbf{Situation Label}              & \textbf{Documents} & \textbf{Sentences} & \textbf{Tokens}  \\ \midrule
\textit{crime and justice}      & 80  & 1209  & 100,635 \\ \midrule
\textit{fire fighting}          & 79   & 1116  & 87,901  \\ \midrule
\textit{technology development} & 81  & 1334  & 122,493 \\ \midrule
\textit{healthcare}             & 81  & 1259  & 117,962 \\ \midrule
\textit{economy}                & 78   & 1264   & 110,605  \\ \bottomrule
\end{tabular}
\endgroup
\caption{\textbf{Data Statistics:} Situation-specific news reports are sampled from GDELT.
Each document (or article) is split into short contexts $C_1, \dots, C_N$ (of $3$ sentences) before being passed into the Operator to generate the sematic representation.  
}
\label{tab:operator_data}
\end{table}

%% file: Tables/operatorframeworks.tex
\begin{table*}[t]
\centering
\small
\begin{tabular}{lccc}
\toprule
\multirow{2}{*}{Situation} & \multicolumn{3}{c}{Ours vs. Baseline} \\
\cmidrule(lr){2-4}
& vs. Zhan et al. & vs. Shi \& Lin & vs. Chanin \\
& (GLEN) & (BERT-SRL) & (FST) \\
\midrule
Crime \& Justice & 0.90 (0.10) & 0.96 (0.04) & 0.70 (0.30) \\
Economy & 0.98 (0.02) & 0.96 (0.04) & 0.86 (0.14) \\
Firefighting & 0.98 (0.02) & 0.98 (0.02) & 0.79 (0.21) \\
Healthcare & 1.00 (0.00) & 0.96 (0.04) & 0.94 (0.06) \\
Tech. Development & 0.96 (0.04) & 0.96 (0.04) & 0.86 (0.14) \\
\bottomrule
\end{tabular}
\caption{\textbf{Operator Evaluations}: Comparison with Existing Frameworks: Given a short context, English-speaking annotators are shown the unlabeled outputs of the Operator and a baseline framework (GLEN, BERT-SRL, FST) and asked to select the one which best summarizes the semantics in the text. The Operator is preferred over baselines across situations.}
\label{tab:operator-framework}
\end{table*}